%% file: main.tex
\documentclass{article}

\usepackage[preprint]{neurips_2024}
\usepackage[shortlabels]{enumitem}

\usepackage[utf8]{inputenc} %
\usepackage[T1]{fontenc}    %
\usepackage{hyperref}       %
\usepackage{url}            %
\usepackage{booktabs}       %
\usepackage{amsfonts, amsmath, amssymb, mathrsfs, mathtools}       %
\usepackage{nicefrac}       %
\usepackage{microtype}      %
\usepackage{xcolor}         %
\usepackage{csquotes}       %
\usepackage{graphicx}
\usepackage{svg}
\usepackage{subcaption}
\usepackage[most]{tcolorbox}
\usepackage{soul}
\usepackage{listings}
\usepackage{xspace}
\usepackage{xcolor}
\usepackage{colortbl}
\usepackage[font=footnotesize]{caption}
\usepackage{bbm}
\usepackage[normalem]{ulem} %
\usepackage{ragged2e}

\tcbuselibrary{breakable, skins}
\usepackage{spverbatim}

\hypersetup{
    colorlinks=true,
    linkcolor=blue,
    citecolor=blue,
    urlcolor=blue,
}

\newcommand{\nocontentsline}[3]{}
\let\oldaddcontentsline\addcontentsline
\newcommand{\tocless}[2]{%
  \let\addcontentsline\nocontentsline
  #1{#2}
  \let\addcontentsline\oldaddcontentsline}

\graphicspath{{figures/}}

\title{Interpretability in Parameter Space: Minimizing Mechanistic Description Length with Attribution-based Parameter Decomposition}

\author{
  Dan Braun\thanks{Core research contributor}\quad
  Lucius Bushnaq\footnotemark[1]\quad
  Stefan Heimersheim\footnotemark[1]\quad
  Jake Mendel\quad \AND
  Lee Sharkey\thanks{Correspondence to lee@apolloresearch.ai }\quad
\AND 
\textmd{Apollo Research}\thanks{Contributions statement below}
}

\begin{document}

\maketitle
\begin{abstract}

Mechanistic interpretability aims to understand the internal mechanisms learned by neural networks. Despite recent progress toward this goal, it remains unclear how best to decompose neural network parameters into mechanistic components.We introduce \textit{Attribution-based Parameter Decomposition} (APD), a method that directly decomposes a neural network's parameters into components that (i) are faithful to the parameters of the original network, (ii) require a minimal number of components to process any input, and (iii) are maximally simple. Our approach thus optimizes for a minimal length description of the network's mechanisms. We demonstrate APD's effectiveness by successfully identifying ground truth mechanisms in multiple toy experimental settings: Recovering features from superposition; separating compressed computations; and identifying cross-layer distributed representations. While challenges remain to scaling APD to non-toy models, our results suggest solutions to several open problems in mechanistic interpretability, including identifying minimal circuits in superposition, offering a conceptual foundation for `features', and providing an architecture-agnostic framework for neural network decomposition.

\end{abstract}
\abstractpagebreak

\newcounter{boxnumber}
\newcounter{prompt}

\pagebreak

\tableofcontents

\pagebreak

\input{1_Introduction}

\input{2_Method}

\input{3_Results}

\input{5_Discussion}

\begin{raggedright}
\bibliography{references}
\bibliographystyle{plainnat}
\end{raggedright}
\newpage
\appendix
\input{6_appendix}
\end{document}

%% file: 1_Introduction.tex
\setcounter{footnote}{0}

\section{Introduction}\label{sec:intro}
Mechanistic interpretability aims to improve the trustworthiness of increasingly capable AI systems by making it possible to understand their internals. The field's ultimate goal is to map the parameters of neural networks to human-understandable algorithms. A major barrier to achieving this goal is that it is unclear how best to decompose neural networks into the individual mechanisms that make up these algorithms, if such mechanisms exist \citep{bussman2024metasaes, sharkey2025openproblemsmechanisticinterpretability}. This is because the mechanistic components of neural networks do not in general map neatly onto individual architectural components, such as individual neurons \citep{hinton1981parallel, churchland2007temporal, Nguyen_2016_FeatureVisualization}, attention heads \citep{Janiak_Mathwin_Heimersheim_2023, jermyn2023attentionheadsuperposition}, or layers \citep{yun2021sparse, lindsey2024crosscoders, meng2023masseditingmemorytransformer}.

Sparse dictionary learning is currently the most popular approach to tackling this problem \citep{lee2007sparse,yun2021sparse,Sharkey_Braun_Millidge_2022,cunningham2023sparse,bricken2023monosemanticity}.
This method decomposes the neural activations of the model at different hidden layers into sets of sparsely activating latent directions. Then, the goal is to understand how these latent directions interact with the network's parameters to form circuits (or `mechanisms') that compute the activations at subsequent layers \citep{cammarata2020curve, olah2023weightsuperposition, sharkey2024sparsifyagenda,  olah2024hurdles}. However, sparse dictionary learning appears not to identify canonical units of analysis for interpretability \citep{bussman2024metasaes}; suffers from significant reconstruction errors \citep{makelov2024principled, gao2024scalingevaluatingsparseautoencoders}; optimizes for sparsity, which may not be a sound proxy for interpretability in the limit \citep{chanin2024absorptionstudyingfeaturesplitting, till2024truefeatures, ayonrinde2024interpretabilitycompressionreconsideringsae}; and leaves feature geometry unexplained \citep{engels2024languagemodelfeatureslinear, Mendel_2024}, among a range of other issues (see \cite{sharkey2025openproblemsmechanisticinterpretability} for review). These issues make it unclear how to use sparsely activating directions in activation space to identify the network's underlying mechanisms. 
Here, we investigate an approach to more directly decompose neural networks parameters into individual mechanisms.

There are many potential ways to decompose neural network parameters, but not all of them are equally desirable for mechanistic interpretability. For example, a neuron-by-neuron description of how a neural network transforms inputs to outputs is a perfectly accurate account of the network's behavior. But this description would be unnecessarily long and would use polysemantic components. This decomposition fails to carve the network at its joints because it does not reflect the network's deeper underlying mechanistic structure.

We therefore ask what properties an ideal mechanistic decomposition of a neural network's parameters should have. Motivated by the minimum description length principle, which states that the shortest description of the data is the best one, we identify three desirable properties:
\begin{itemize}
    \item {\textbf{Faithfulness}: The decomposition should identify a set of components that sum to the parameters of the original network.\footnote{Faithfulness to the original network's parameters is subtly different from the `behavioral faithfulness of a circuit', which has been studied in other literature \citep{wang2022interpretability}. Components that sum to the parameters of the original network will necessarily exhibit behavior that is faithful to the original network (assuming all components are included in the sum). But behavioral faithfulness does not imply faithfulness to a network's parameters, since different parameters may exhibit the same behavior. Our criterion is therefore stricter, and relates to a decomposition of parameters rather than \cite{wang2022interpretability}'s definition of a circuit.}}
    \item{\textbf{Minimality}: The decomposition should use as few components as possible to replicate the network's behavior on its training distribution. }
    \item{\textbf{Simplicity}: Components should each involve as little computational machinery as possible.} 
\end{itemize}

\begin{figure}[h!] 
    \centering
    \includegraphics[width=1.0\linewidth]{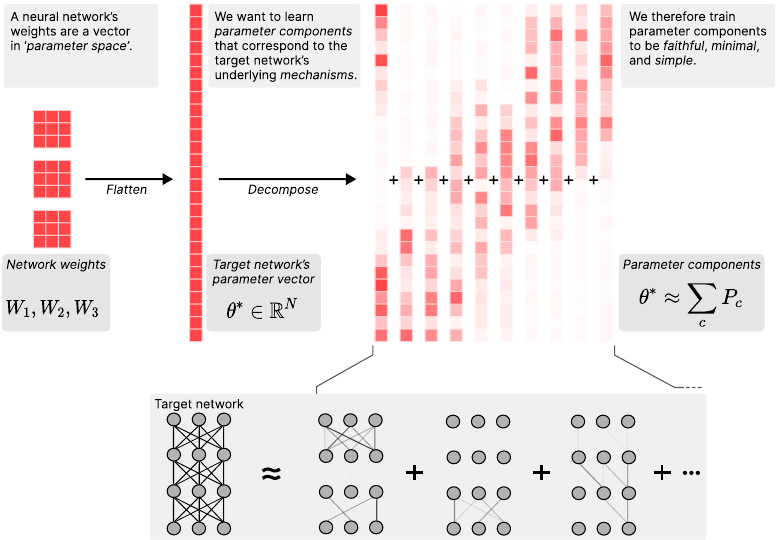}
    \caption{Decomposing a target network's parameters into parameter components that are faithful, minimal, and simple.}\label{fig:spd_overview}
\end{figure}

Insofar as we can decompose a neural network's parameters into components that exhibit these properties, we think it would be justified to say that we have identified the network's underlying \textit{mechanisms}: Faithfulness ensures that the decomposition reflects the parameters of and computations implemented by the network. Minimality ensures the decomposition comprises specialised components that that play distinct roles. And simplicity encourages the components to be individual, basic computational units, rather than compositions of them.

To this end, we introduce \textit{Attribution-based Parameter Decomposition (APD)}, a method that decomposes neural network parameters into components that are optimized for these three properties. In brief, APD involves decomposing the parameter vector of any neural network into a sum of \textit{parameter components}. They are optimized such that they sum to the target parameters while only a minimal number of them are necessary for the causal process that computes the network's output for any given input. They are also optimized to be less complex individually than the entire network, in that they span as few directions in activation space as possible across all layers. APD can be understood as an instance of a broader class of \textit{Linear Parameter Decomposition} (LPD) methods, a term we attempt to make precise in this paper.

Our approach leverages the idea that, for any given input, a neural network should not require all of its mechanisms simultaneously \citep{veit2016residual, zhang2022moefication, dong2023attention}.  On any given input, it should be possible to ablate unused mechanisms without influencing the network's computations. This would let us study the mechanisms in relative isolation, making them easier to understand.
For example, suppose a neural network uses only one mechanism to store the knowledge that `\textit{The sky is blue}' in its parameters. Being a `mechanism', as defined above, it is maximally simple, but may nevertheless be implemented using multiple neurons scattered over multiple layers of the model. Despite being spread throughout the network, we contend that there is a single vector in parameter space that implements this knowledge. On inputs where the model uses this stored knowledge, the model's parameters along this direction cannot be varied without changing the model's output. But on inputs where the model does not use this fact, ablating the model parameters along this direction to zero should not change the output.

Our method has several connections to other contemporary approaches in mechanistic interpretability, such as sparse dictionary learning \citep{Sharkey_Braun_Millidge_2022, cunningham2023sparse, bricken2023monosemanticity, braun2024identifying, dunefsky2024transcodersinterpretablellmfeature, ayonrinde2024interpretabilitycompressionreconsideringsae, lindsey2024crosscoders}, causal mediation analysis \citep{vig2020causal, wang2022interpretability, conmy2024towards, syed2023attributionpatchingoutperformsautomated, kramár2024atpefficientscalablemethod, geiger2024findingalignmentsinterpretablecausal}, weight masking \citep{mozer1988skeletonization, phillips2019explanatorymasksneuralnetwork, csordás2021neuralnetsmodularinspecting, decao2021sparseinterventionslanguagemodels}, and others, while attempting to address many of their shortcomings. It also builds on work that explores the theory of computation in superposition \citep{Vaintrob_Mendel_Kaarel_2024, Bushnaq_Mendel_2024}.

This paper is structured as follows: We first describe our method in Section \ref{sec:method}. In Section \ref{sec:results}, we provide empirical support for our theoretical work by applying APD to three toy models where we have access to ground-truth mechanisms. First, in a toy model of superposition, APD recovers mechanisms corresponding to individual input features represented in superposition (Section \ref{sec:TMS_results}). Second, in a model performing compressed computation -- where a model is tasked with computing more nonlinear functions than it has neurons -- APD finds parameter components that represent each individual function (Section \ref{sec:residmlp_1layer}). Third, when extending this model of compressed computation to multiple layers, APD is still able to learn components that represent the individual functions, even those that span multiple layers (Section \ref{sec:residmlp_2layer}).
In Section \ref{sec:discussion}, we discuss our results, the current state of APD, and possible next steps in its development, with conclusions in Section \ref{sec:conclusion}. We include a detailed discussion on related work in Section \ref{sec:related_work}.

%% file: 2_Method.tex
\section{Method: Attribution-based Parameter Decomposition}\label{sec:method}

In this section, we outline our method, Attribution-based Parameter Decomposition (APD). First, we outline why we define `mechanisms' as vectors in parameter space (Section \ref{sec:method_defining_mech_space}). Then, we discuss how our method optimizes parameter components to be faithful, minimal, and simple, thus identifying the network's mechanisms (Section \ref{sec:method_APD_brief_intro}).  While a brief description of APD suffices to understand our experiments, a more detailed description and motivation can be found in Appendix \ref{app:full_spd_explan}.

\subsection{Defining `mechanism space' as parameter space}\label{sec:method_defining_mech_space}

To identify a neural network's mechanisms, we must first identify the space in which they live. The weights of neural networks can be flattened into one large parameter vector in \textit{parameter space} (Figure \ref{fig:spd_overview}). During learning, gradient descent iteratively etches a neural network’s mechanisms into its parameter vector. This makes it natural to look for mechanisms in the same vector space as the whole network. 

Vectors in parameter space also satisfy a broad range of criteria that we require individual mechanisms to have. \textit{Mechanism space} should:
\begin{itemize}
    \item{ \textbf{Span the same functional range as the target network}: We want mechanisms that perform a subcomponent of the algorithm implemented by the target neural network. We therefore expect mechanisms to lie somewhere in between “\textit{Doing everything the target network does}” and “\textit{Doing nothing}”. Parameter space contains such mechanisms: The target network's parameter vector does everything that the target network does. And the zero parameter vector serves as a `null mechanism'. Vectors that lie `in between' serve as candidates for individual mechanisms\footnote{For this to be meaningful, we need a reasonable definition of what it means for vectors to lie `in between' the target network's parameter vector and the zero vector. One reasonable definition is that vectors that are `in between' should have a lower magnitude than the target network's parameter vector and have positive cosine similarity with the target network’s parameters. This definition is implied by the method introduced in this work, although it does not optimize for these properties directly.}. } 

    \item{ \textbf{Accommodate basis-unaligned mechanisms}: It has long been known that neural representations may span multiple neurons \citep{hinton1981parallel, churchland2007temporal, Nguyen_2016_FeatureVisualization}. However, even more recent work suggests that representations may span other architectural components, such as separate attention heads \citep{Janiak_Mathwin_Heimersheim_2023, jermyn2023attentionheadsuperposition} or even layers \citep{yun2021sparse, lindsey2024crosscoders, meng2023masseditingmemorytransformer}. Vectors in parameter space span all of these components and can therefore implement computations that happen to be distributed across them.}

    \item{ \textbf{Accommodate superposition}: Neural networks appear to be able to represent and perform computation on variables in superposition \citep{elhage2022toy, Vaintrob_Mendel_Kaarel_2024, Bushnaq_Mendel_2024}. 
    We would like a space that can compute more functions than they have neurons. Vectors in parameter space support this requirement, theoretically \citep{Bushnaq_Mendel_2024} and in practice, as we will demonstrate in our experiments.}

    \item{ \textbf{Accommodate multidimensional mechanisms}: Some representations in neural networks appear to be multidimensional \citep{engels2024languagemodelfeatureslinear}. We therefore want to be able to identify mechanisms that can do multidimensional computations on these representations. Vectors in parameter space satisfy this requirement.}
\end{itemize}

Having defined mechanism space as parameter space, we now want a method to identify a set of parameter components that correspond to the network's underlying mechanisms. In particular, we want to identify parameter components that satisfy the faithfulness, minimality, and simplicity criteria.

\subsection{Identifying networks' mechanisms using Attribution-based Parameter Decomposition}\label{sec:method_APD_brief_intro}

APD aims to minimize the description length of the mechanistic components used by the network \textit{per data point} over the training dataset. It decomposes the network's parameters $\theta^* \in \mathbb{R}^N$ into a set of parameter components and directly optimizes them to be faithful, minimal, and simple. A discussion of how APD can be understood as an instance of a broader class of `linear parameter decomposition' methods can be found in Appendix \ref{app:LPD}, and more detailed discussion of how APD is based on the Minimum Description Length principle can be found in Appendix \ref{app:MDL}. 

\paragraph{Optimizing for faithfulness:} We decompose a network's parameters $\theta^{*}_{l,i,j}$, where $l$ indexes the network's weight matrices and $i,j$ index rows and columns, by defining a set of $C$ parameter components $P_{c,l,i,j}$. Their sum is trained to minimize the mean squared error (MSE) with respect to the target network's parameters, $\mathcal{L}_{\text{faithfulness}}=\text{MSE}(\theta^*,\sum^C_{c=1} P_c)$.

\paragraph{Optimizing for minimality:}The parameter components are also trained such that, for a given input, a minimal number of them is used to explain the network's output (Figure \ref{fig:minimality_step}). 
To achieve this, we use two steps: 

\begin{figure}[ht!] 
    \centering
    \includegraphics[width=0.8\linewidth]{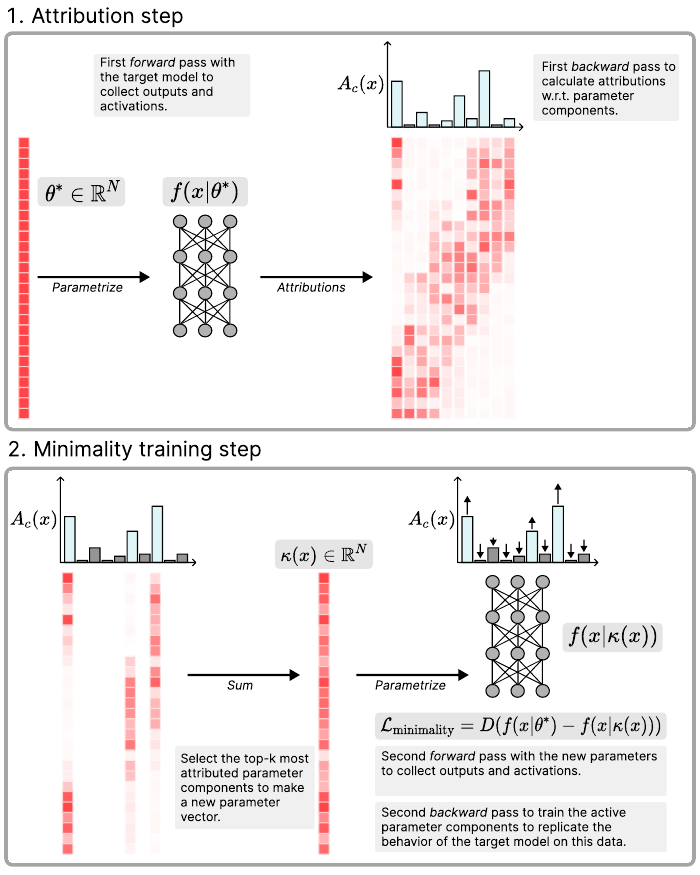}
    \caption{\textbf{Top:} Step 1: Calculating parameter component attributions $A_c (x)$. \textbf{Bottom:} Step 2: Optimizing minimality loss $\mathcal{L}_{\text{minimality}}$.}\label{fig:minimality_step}
\end{figure}

\begin{enumerate}
    \item \textbf{Attribution step:} We want to estimate the causal importance of each parameter component $P_c$ for the network's output on each datapoint $f_{\theta^*}(x)$. In this step, we therefore calculate the \textit{attributions} of each parameter component with respect to the outputs, $A_c(x) \in \mathbb{R}$. 
    It would be infeasibly expensive to compute this exactly, since it would involve a large number of causal interventions, requiring one forward pass for every possible combination of component ablations. We therefore use an approximation. In this work, we use gradient attributions \citep{mozer1988skeletonization, molchanov2017pruningconvolutionalneuralnetworks, nanda2022attribution, syed2023attributionpatchingoutperformsautomated}, but other attribution methods may also work.
    This step therefore involves one forward pass with the target model to calculate the output and one backward pass per output dimension\footnote{In the future, this could possibly be reduced to one backward pass with stochastic sources techniques (See \cite{knechtli2016lattice} (Chapter $3.6$) for an introduction to these methods).} to compute the attributions with respect to the parameters, which are used to calculate the attributions with respect to each parameter component (Equation \ref{eq:attribution}). 
    \item \textbf{Minimality training step}: 
    We sum only the top-$k$ most attributed parameter components, yielding a new parameter vector $\kappa(x) \in \mathbb{R}^N$, and use it to perform a forward pass. We train the output of the top-$k$ most attributed parameter components to match the target network's outputs by minimizing $\mathcal{L}_{\text{minimality}}=D(f_{\theta^*}(x), f_{\kappa(x)}(x))$, where $D$ is some distance or divergence measure (Equation \ref{eq:loss_topk}). This step trains the \textit{active} parameter components to better reconstruct the target network's behavior on a given data point. This should increase the attribution of active components on that data. In some cases, we also train some of the hidden activations to be similar on both forward passes, since it may otherwise be possible for APD to learn solutions that produce the same outputs using different computations (See Appendix \ref{app:top_k} for details).  
\end{enumerate}

In our experiments, we use batch top-$k$ \citep{bussmann2024batchtopk} to select a fixed number of active parameter components for the minimality training step (a.k.a sparse forward pass) per batch. This sidesteps the issue of needing to select a specific number of active parameter components for each sample, although does present other issues (see Appendix \ref{app:resid_mlp_1layer_apd}).

\paragraph{Optimizing for simplicity:} The components are also trained to be `simpler' than the parameters of the target network. 
We would like to penalize parameter components that span more ranks or more layers than necessary by minimizing the sum of the ranks of all the matrices in active components: $\sum^C_{c=1} s_c(x)\sum_l\text{rank}(P_{c,l})$, where $s_c(x)\in\{0,1\}$ indicates active components. 
In practice, we minimize the $L_p$ norm of the singular values of weight matrices in active components using a loss $\mathcal{L}_{\text{simplicity}}(x)=\sum^C_{c=1} s_c(x)\sum_{l,m}\vert\vert \lambda_{c,l}\vert\vert^p_p$, where $\lambda_{c,l,m}$ are the singular values of parameter component $c$ in layer $l$. This is also known as the Schatten-$p$ norm\footnote{Since $p\in(0,1)$, the Schatten-$p$ norm and $L_p$ norms here are technically quasi-norms. For brevity, we refer to them as norms throughout.}.
For a discussion of how to calculate Schatten norms efficiently, see Appendix \ref{app:top_k}.
\paragraph{Biases} Currently, we do not decompose the network's biases. Biases can be folded into the weights by treating them as an additional column in each weight matrix, meaning they can in theory be decomposed like any other type of parameter. However, in this work, for simplicity we treat them as their own parameter component that is active for every input, and leave their decomposition for future work. 

\paragraph{Summary:} In total, we use three losses: 
\begin{enumerate}
    \item{A faithfulness loss ($\mathcal{L}_{\text{faithfulness}}$), which trains the sum of the parameter components to approximate the parameters of the target network.}
    \item{A minimality loss ($\mathcal{L}_{\text{minimality}}$), which trains the top-$k$ most attributed parameter components on any given input to produce the same output (and some of the same hidden activations) as the target network, thereby increasing their attributions on those inputs.}
    \item{A simplicity loss ($\mathcal{L}_{\text{simplicity}}$), which penalizes parameter components that span more ranks or more layers than necessary.}
\end{enumerate}

%% file: 3_Results.tex
\section{Experiments: Decomposing neural networks into mechanisms using APD}\label{sec:results}

In this section, we demonstrate that APD succeeds at finding faithful, minimal, and simple parameter components in three toy settings with known ‘ground truth mechanisms’. These are 
\begin{enumerate}
    \item \citet{elhage2022toy}'s toy model of superposition (Section \ref{sec:TMS_results});
    \item A novel toy model of compressed computation, which is a model that computes more nonlinear functions than it has neurons (Section \ref{sec:residmlp_1layer});
    \item A novel toy model of cross-layer distributed representations (Section \ref{sec:residmlp_2layer}).
\end{enumerate}
In all three cases, APD successfully identifies the ground truth mechanisms up to a small error. The target models are trained using AdamW \citep{loshchilov2019adamw}, though we also study a handcoded model in Appendix \ref{app:handcoded_gated_model}. Additional figures and training logs can be found \href{https://api.wandb.ai/links/apollo-interp/j93iqupv}{here}. All experiments were run using \href{https://github.com/ApolloResearch/apd}{github.com/ApolloResearch/apd}. Training details and hyperparameters can be found in Appendix \ref{app:training-details}.

\subsection{Toy Model of Superposition}\label{sec:TMS_results}
\begin{figure}
    \centering
    \includegraphics[width=1\linewidth]{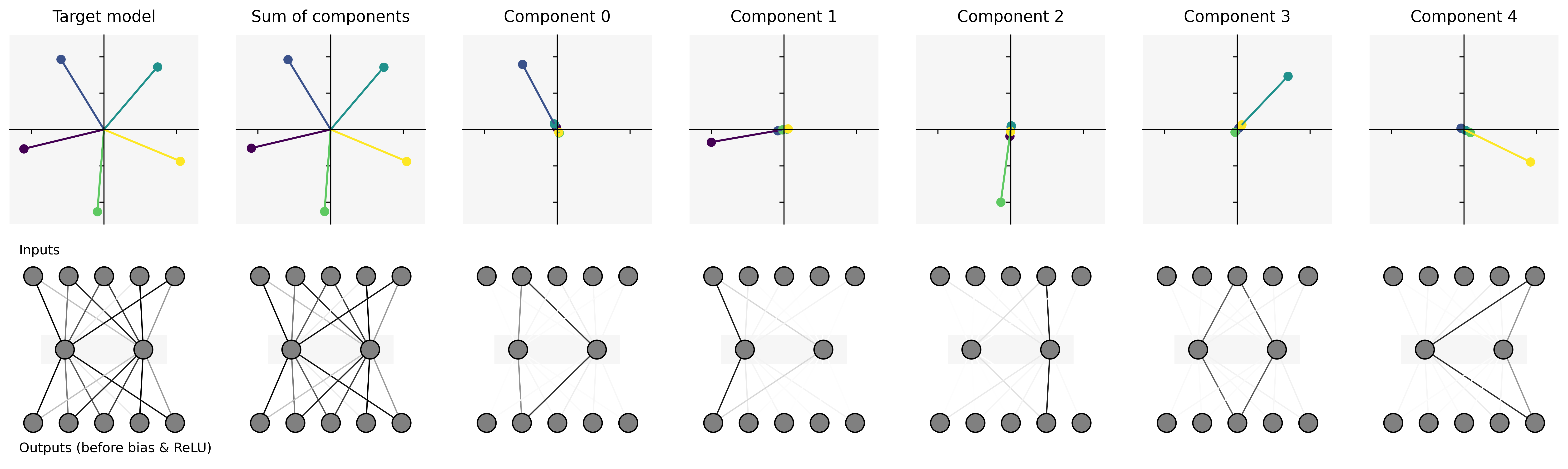}
    \caption{Results of running APD on TMS. \textbf{Top row:} Plot of the columns of the weight matrix of the target model, the sum of the APD parameter components, and each individual parameter component. Each parameter component corresponds to one mechanism, which in this model each correspond to one `feature' in activation space \citep{elhage2022toy}. \textbf{Bottom row:} Depiction of the corresponding parametrized networks.}
    \label{fig:tms-combined}
\end{figure}

Our first model is \cite{elhage2022toy}'s toy model of superposition (TMS), which can be written as $\hat{x}= \text{ReLU}(W^\top W x + b)$, with weight matrix $W \in \mathbb{R}^{m_1 \times m_2}$.
The model is trained to reconstruct its inputs, which are sparse sums of one-hot $m_2$-dimensional input features, scaled to a random uniform distribution $[0,1]$. 
Typically, $m_1 < m_2$, so the model is forced to `squeeze' representations through a $m_1$-dimensional bottleneck. When the model is trained on sufficiently sparse data distributions, it can learn to represent features in superposition in this bottleneck. For certain values of $m_1$ and $m_2$, the columns of the W matrix often form regular polygons in the $m_1$-dimensional hidden activation space (Figure \ref{fig:tms-combined} leftmost panel).

What are the `ground truth mechanisms' in this toy model? Let us define a set of matrices $\{Z^{(c)}\}$ that are zero everywhere except in the $c^{\text{th}}$ column, where they take the values $W_{:, c}$:
\begin{equation}
   Z^{(c)}_{:,j} = 
   \begin{cases} 
   W_{:,c} & \text{if } j = c, \\ 
   \textbf{0} & \text{otherwise}.
   \end{cases}
\end{equation}
The data are sparse, so only some of the model's weights $W$ are used on any given datapoint. Suppose we have a datapoint where only dataset feature $c$ is active. On this datapoint, we can replace $W$ with $Z^{(c)}$ and the model outputs would be almost identical (since the interference terms from inactive features should be small and below the learned $\text{ReLU}$ threshold). Intuitively, a column of $W$ is only `used' if the corresponding data feature is active. This makes the matrices $\{Z^{(c)}\}$ good candidates for optimal `minimality'. 
The matrices are also `faithful', since $\sum_c Z^{(c)} = W$. 
They are also very simple because each matrix is rank 1, consisting of the outer product of the column of $W_{:,c}$ and the one-hot vector $e_c \in \mathbb{R}^{m_2}$ that indexes the nonzero column $c$:
\begin{equation}
   Z^{(c)}=  W_{:,c} e_c^\top
\end{equation}

The matrices $\{Z^{(c)}\}$ are therefore reasonable candidates for the ground truth `mechanisms' of this model \footnote{The $c^{\text{th}}$`mechanism' in this model technically corresponds to $Z^{(c)}$ \textit{and} the $c^{\text{th}}$ element of the bias. For simplicity, we do not decompose biases in our current implementation and treat all biases as one component that is always active.}. We would therefore like APD to learn parameter components that correspond to them.

\subsubsection*{APD Results: Toy Model of Superposition}

We find that APD can successfully learn parameter components $\{P_c\}$ that closely correspond to the matrices $\{Z^{(c)}\}$ (Figure \ref{fig:tms-combined}). We observe that the sum of the components is equal to $W$ in the target network.

For illustrative purposes, we have focused on the setting with $5$ input features ($m_2=5$) and a hidden dimension of $2$ ($m_1=2$). However, training an APD model (and to a lesser extent, a target model) in this setting is very brittle and less effective than a higher-dimensional setting. Indeed, for the $2$-dimensional hidden space setting, the results presented in this section required an adjustment of using attributions taken from the APD model rather than the target model (an adjustment that proved not to be beneficial in other settings). We expect that this brittleness has to do with the large amount of interference noise between the input features when projected onto the small $2$-dimensional space. We thus also analyze a setting with $40$ input features and $10$ hidden dimensions. We use $\text{TMS}_{5-2}$ to denote the setting with $5$ input features and $2$ hidden dimensions, and $\text{TMS}_{40-10}$ to denote the setting with $40$ input features and $10$ hidden dimensions.

To show how close the learned parameter components are to the columns of $W$ in the target model, we measure the angle between each column of $W$ and the corresponding column in the component it lines up best with. We also measure how close their magnitudes are.  
To quantify the angles, we calculate the mean max cosine similarity (MMCS) \citep{Sharkey_Braun_Millidge_2022} %
\begin{equation}
\text{MMCS}(W, \{P_c\}) = \frac{1}{m_2}\sum_{j=1}^{m_2}\max_c(\frac{P_{c,:,j}\cdot W_{:,j}}{\vert\vert P_{c,:,j}\vert\vert_2 \vert\vert W_{:,j}\vert\vert_2})\,,
\end{equation}
where $c\in C$ are parameter component indices and $j\in[1,m_2]$ are input feature indices. 
A value of $1$ for MMCS indicates that, for all input feature directions in the target model, there exists a parameter component whose corresponding column points in the same direction. 
To quantify how close their magnitudes are, we calculate the mean L2 Ratio (ML2R) between the Euclidean norm of the columns of $W$ and the Euclidean norm of the columns of the parameter components $P_c$ with which they have the highest cosine similarity
\begin{equation}
\text{ML2R}(W, \{P_c\}) = \frac{1}{m_2}\sum_{j=1}^{m_2} \frac{\vert\vert P_{\text{mcs}(j),:,j}\vert\vert_2}{\vert\vert W_{:,j}\vert\vert_2}\,,
\end{equation}
where $\text{mcs}(j)$ is the index of the component that has maximum cosine similarity with weight column $j$ of the target model.
A value close to $1$ for the ML2R indicates that the magnitude of each parameter component is close to that of its corresponding target model column.

The MMCS and ML2R for both $\text{TMS}_{5-2}$ and $\text{TMS}_{40-10}$ are shown in Table \ref{tab:tms_mmcs}. We see in both settings that the MMCS values are $\approx 1$. 
This indicates that the parameter components are close representations of the target model geometrically.

However, the ML2R is close to $0.9$, implying there is some amount of `shrinkage', reminiscent of feature shrinkage in SAEs \citep{jermyn2024tanh, wright2024suppression}. 
We speculate that shrinkage in APD is caused by a forced trade-off between top-$k$ reconstruction $\mathcal{L}_{\text{minimal}}$ and the Schatten norm penalty $\mathcal{L}_{\text{simplicity}}$. 
In this specific case, we suspect it might be due to noise in the target model output due to high interference between the input features. The parameter components are incentivised by $\mathcal{L}_{\text{minimal}}$ to reconstruct this noise such that each learns small amounts of different ground truth mechanisms $\{Z^{(c)}\}$.  Additional visualizations of the TMS APD models can be found in a WandB report \href{https://api.wandb.ai/links/apollo-interp/j93iqupv}{here}.

\begin{table}[t]
\centering
\begin{tabular}{lccc}
\toprule
& MMCS & ML2R \\
\midrule
$TMS_{5-2}$ & 0.998 ± 0.000 & 0.893 ± 0.004  \\
$TMS_{40-10}$ & 0.996 ± 0.003 & 0.935 ± 0.001  \\
\bottomrule
\end{tabular}
\vskip 0.15in
\caption{Mean max cosine similarity (MMCS) and mean L2 ratio (ML2R) with their standard deviations (to $3$ decimal places) between learned parameter components and target model weights for TMS. The MMCS is very close to $1.0$, indicating that every column in the target model has a corresponding column in one of the components that points in almost the same direction. The ML2R is below $1.0$, indicating some amount of shrinkage in the components compared to the original model.}

\label{tab:tms_mmcs}
\end{table}

\begin{figure}
\centering
\includegraphics[width=1.1\linewidth]{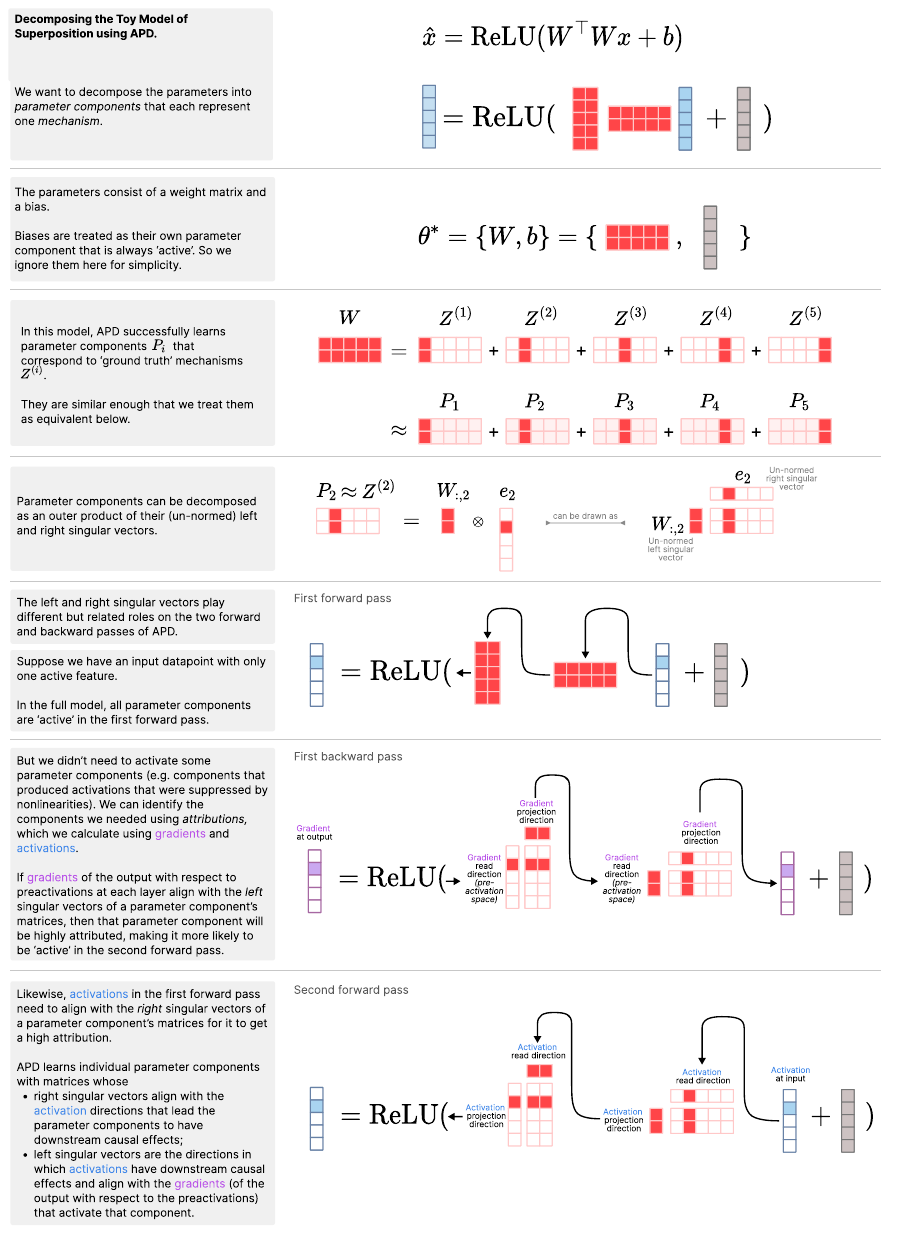}
    \caption{Decomposing TMS with APD.}
    \label{fig:apd-svd}
\end{figure} 

It is worth reflecting on the differences between the APD solution and the decompositions that other commonly used matrix decomposition methods would yield, such as singular value decomposition \citep{millidge2022svd, meller2023svr} or non-negative matrix factorization \citep{petrov2021weight, voss2021visualizing}. Those methods can find at most $\text{rank}(W)=\text{min}(m_1, m_2)$ components, and therefore could not decompose $W$ into its ground truth mechanisms even in principle. 

The toy model studied in this section was initially developed in order to demonstrate that neural networks can represent variables `in superposition' using an overcomplete basis of the bottleneck activation space. However, our work decomposes the model, not activation space. Nevertheless, the mechanisms identified by our method \textit{imply} an overcomplete basis in the activation space: The rank 1 mechanisms $Z^{(c)}$ can be expressed as an outer product of their (un-normed) left and right singular vectors $W_{:,c} e_c^\top$. The left singular vectors (corresponding to the columns of $W$) are an overcomplete basis of the $m_1$-dimensional hidden activation space\footnote{Equivalently, since we use $W$ and $W^\top$ for the matrices in this model, the right hand components of $e_c W_{:,c}^\top$ also imply an overcomplete basis of the $m_1$-dimensional hidden activation space.}. Parameter vectors can thus imply overcomplete bases for the activation spaces that they interact with, even though they do not form an overcomplete basis for parameter space.

The structure of this matrix decomposition is also revealing: We can think of $P_c$ as `reading' from the $e_c^\top$ direction in the input space and projecting to the $W_{:,c}$ direction in the bottleneck activation space (Figure \ref{fig:apd-svd}). Since we use $W$ and $W^\top$ in this model, in the next layer we can also think of this parameter component `reading' from the $W_{:,c}^\top$ in the bottleneck activation space and projecting to the $e_c$ direction in the pre-ReLU activation space. In the backward pass, the roles are reversed: Directions that were `reading' directions for activations become `projecting' directions for gradients, and vice versa. In general, networks will learn parameter components consisting of matrices whose right singular vectors align with the hidden activations on the forward pass and whose left singular vectors align with gradients of the output with respect to the preactivations on the backward pass. 
Thus, they will ignore directions along which there are no activations, as well as directions that have no downstream causal effects.

\subsection{Toy Model of Compressed Computation}
\label{sec:residmlp_1layer}

While the previous example (TMS) analyzed APD on a model that stored more features than dimensions, here we examine APD on a model performing more computations than it has neurons — a phenomenon that we term \textit{compressed computation}. We chose this model because neural networks trained on realistic tasks may often perform more computations than they have neurons. Compressed computation is very similar to the \enquote{Computation in Superposition} toy model introduced by \cite{elhage2022toy}, but our architecture and task differ. 
A key characteristic of representation in superposition \citep{elhage2022toy} and computation in superposition \citep{Bushnaq_Mendel_2024} is a dependence on input sparsity. 
We suspect our model's solutions to this task might not depend on the sparsity of inputs as much as would be expected, potentially making `compressed computation' and `computation in superposition' subtly distinct phenomena. 
But we could not conclusively establish that distinction, since experiments investigating it transpired to be more complicated than they initially appeared. We leave a more detailed study of this distinction for future work. To avoid potential confusion, we opted for a distinct term. 

We train a target network to approximate a function of sparsely activating inputs $x_i\in[-1, 1]$, using a Mean Squared Error (MSE) loss between the model output and the labels. The labels we train the model to predict are produced by the function $y_i = x_i + \text{ReLU}(x_i)$. Crucially, the task involves learning to compute more ReLU functions than the network has neurons.

\begin{figure}
    \centering
    \includegraphics[width=0.5\linewidth]{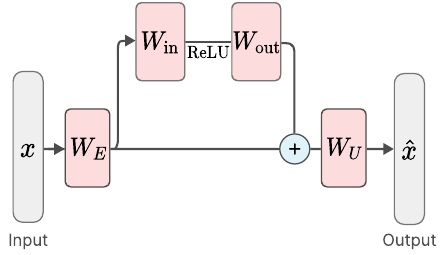}
    \caption{The architecture of our Toy Model of Compressed Computation using a $1$-layer residual MLP. We fix $W_E$ to be a randomly generated matrix with unit norm rows, and $W_U={W_E}^\top$.}
    \label{fig:resid_mlp_1l_architecture}
\end{figure}

The target network is a residual MLP, consisting of a residual stream width of $d_{\rm resid}=1000$, a single MLP layer of width $d_{\rm mlp}=50$, a fixed, random embedding matrix with unit norm rows $W_E$, an unembedding matrix $W_U=W_E^\top$, and $100$ input features. See Figure \ref{fig:resid_mlp_1l_architecture} for an illustration of the network architecture. The large residual stream $d_{\rm resid}=1000$ was chosen as the trained target network performed better than the naive monosemantic baseline in this setting (small values of $d_{\rm resid}$ lead to higher interference and thus worse model performance). We chose fixed, instead of trained, embedding matrices to make it simpler to calculate the optimal monosemantic baseline and to simplify training.

A naive solution to this task is to dedicate one neuron each to the computation of the first $d_{\rm mlp}$ functions, and to ignore the rest. This monosemantic baseline solution would perform perfectly for inputs that contained active features in only the first $d_{\rm mlp}$ input feature indices but poorly for all other inputs.

To understand how each neuron participates in computing the output for a given input feature, we measure what we call the neuron's \textit{contribution} to each input feature computation. For each neuron, this contribution is calculated by multiplying two terms:
\begin{enumerate}
    \item How strongly the neuron reads from input feature $i$ (given by $W_\text{IN} {W_E}_{[:,i]}$).
    \item How strongly the neuron's output influences the model's output for index $i$ (given by ${W_U}_{[i,:]} W_\text{OUT}$).
\end{enumerate}

Mathematically, we compute neuron contributions for each input feature computation $i\in[0,99]$ by $({W_U}_{[i,:]} W_\text{OUT}) \odot (W_\text{IN} {W_E}_{[:,i]})$, where $\odot$ denotes element-wise multiplication. A large positive contribution indicates that the neuron plays an important role in computing the output for input feature $i$. Figure \ref{fig:resid-mlp-weights} (top) shows the neurons involved in the computation of the first $10$ input features of the target model and their corresponding contribution values. We analyze this target model in more detail in Appendix \ref{app:resid_mlp_target}.

The goal for APD in this setting is to learn parameter components that correspond to the computation of each input feature in the target model, despite these computations involving neurons that are used to compute multiple input features. For simplicity, we only decompose the MLP weights and do not decompose the target model's embedding matrix, unembedding matrix, or biases.
 
We found that parameter components often `die' during training, such that no input from the training dataset can activate them. For this reason, we train with $130$ parameter components. This gives APD a better chance of learning all $100$ of the desired parameter components corresponding to unique input feature computations. More APD training details are given in Appendix \ref{app:training-details}.

\subsubsection*{APD Results: Toy Model of Compressed Computation}
\begin{figure}
    \centering
    \includegraphics[width=1\linewidth]{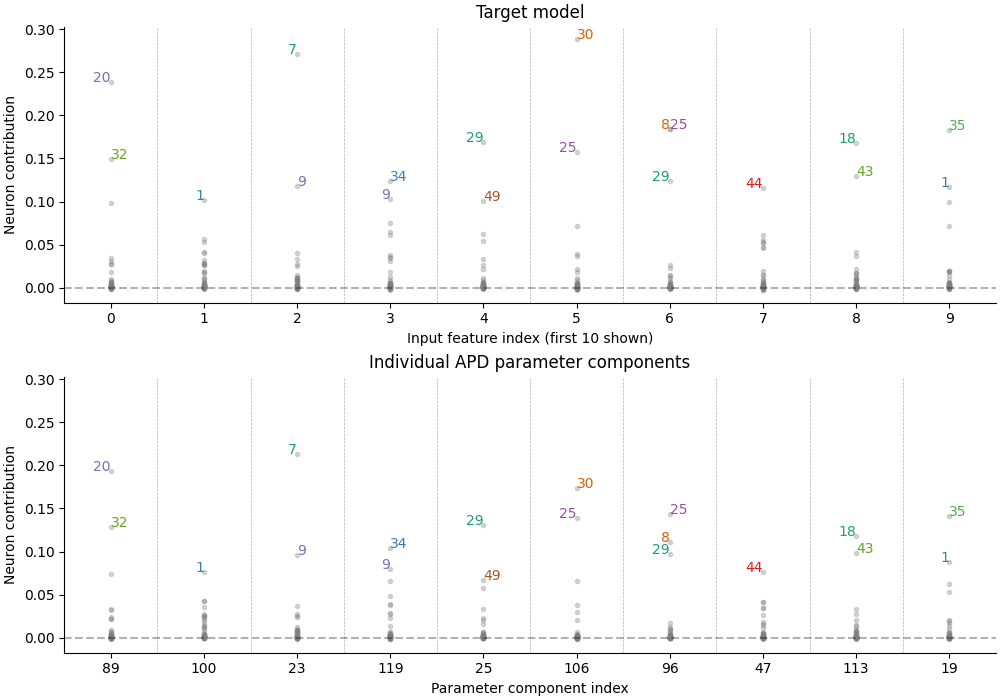}
    \caption{Similarity between target model weights and APD model components for the first $10$ (out of $100$) input feature dimensions. \textbf{Top}: Neuron contributions measured by $({W_U}_{[i,:]} W_\text{OUT}) \odot (W_\text{IN} {W_E}_{[:,i]})$ for each input feature index $i\in[0,9]$, where $\odot$ is an element-wise product.  \textbf{Bottom}: Neuron contributions for the predominant parameter components, measured by $\max_k [({W_U}_{[i,:]} {W_\text{OUT}}_k) \odot ({W_\text{IN}}_k {W_E}_{[:,i]})]$ for each feature index $i\in[0,9]$. The neurons are numbered from $0$ to $49$ based on their raw position in the MLP layer. An extended version of this figure showing all input features and parameter components can be found \href{https://api.wandb.ai/links/apollo-interp/h5ekyxm7}{here}.}
    \label{fig:resid-mlp-weights}
\end{figure}

Despite the target model computing more functions ($100$) than it has neurons ($50$), we find that APD can indeed learn parameter components that each implement $y_i = x_i + \text{ReLU}(x_i)$ for unique input dimensions $i\in \{0,\cdots,99 \}$. Figure \ref{fig:resid-mlp-weights} provides a visual representation of a set of learned parameter components. It shows how the computation that occurs for each input feature in the target network (top) is well replicated by individual parameter components in the APD model (bottom). We see that, for each input feature, there is a corresponding parameter component that uses the same neurons to compute the function as the target model does. 
Note that while we do not see a perfect match between the target model and the APD model, a perfect match would not actually be expected nor desirable: the neuron contribution scores of the target model can contain interference terms from the overlapping mechanisms of other features, which a single APD parameter component is likely to filter out. 
However, there is some `shrinkage', similar to what we observe in the results on the TMS model (see Section \ref{sec:TMS_results}). 
Here, much of the shrinkage is due to batch top-$k$ forcing APD on some batches to activate more components than there are features in the input, thereby spreading out input feature computations across multiple components. We discuss some of the trade-offs when setting batch top-$k$ in Appendix \ref{app:resid_mlp_1layer_apd}.

\begin{figure}
    \centering
    \includegraphics[width=1\linewidth]{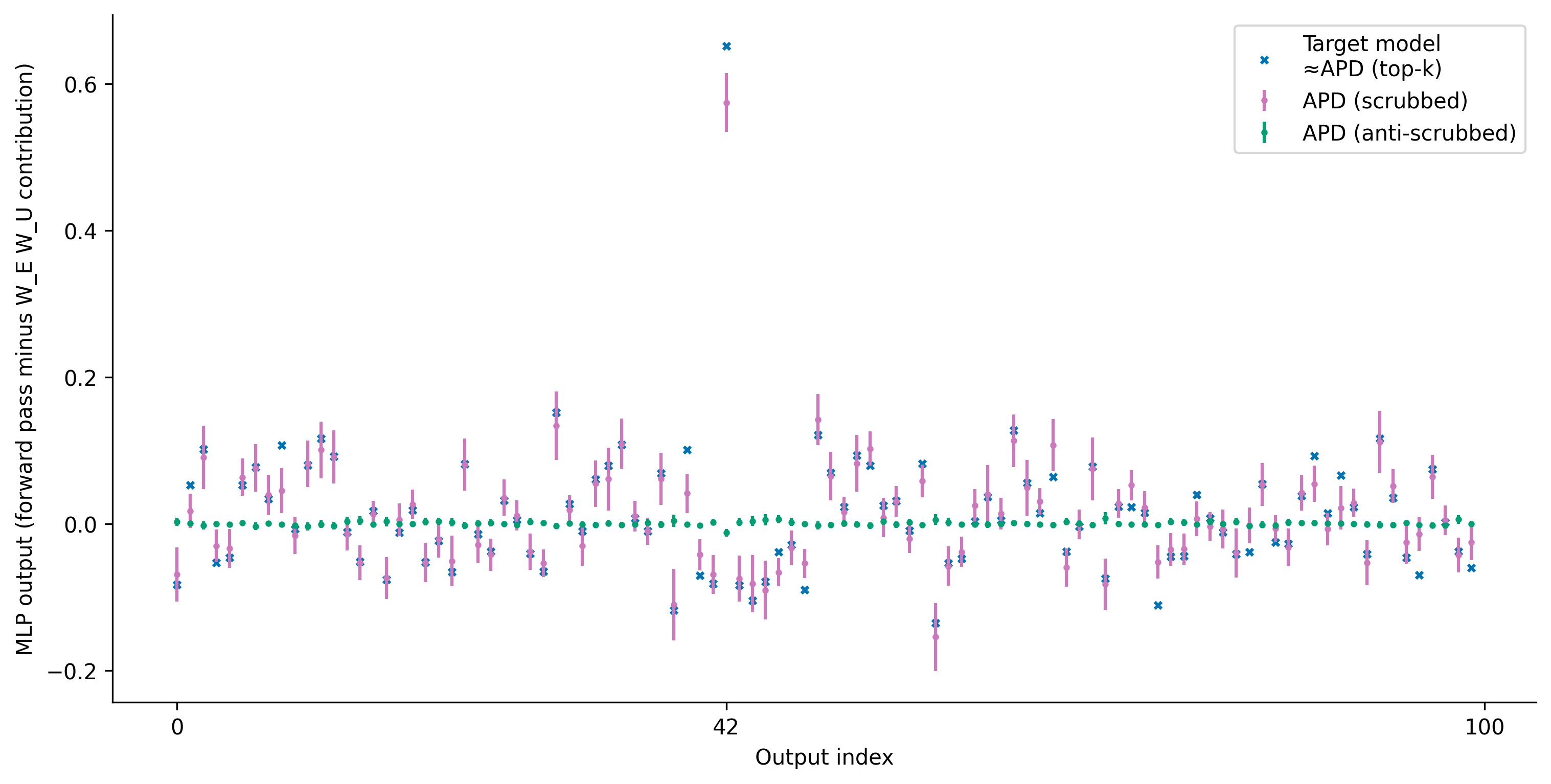}
    \caption{Output of multiple APD forward passes with one-hot input $x_{42} = 1$ over $10$k samples, where half of the parameter components are ablated in each run. Purple lines show `scrubbed' runs (parameter component corresponding to input index $42$ is preserved), while green lines show `anti-scrubbed' runs (component $42$ is among those ablated). The target model output is shown in blue, which is almost identical to the output on the APD sparse forward pass (i.e. APD (top-$k$)). In this plot we only show the MLP output for clearer visualization. The embedding matrices are not decomposed and thus the residual stream contribution does not depend on APD components.}
    \label{fig:weight-linearity}
\end{figure}

Next, we investigate whether individual APD components have minimal influence on forward passes where their corresponding input feature is not active using a Causal Scrubbing-inspired experiment \citep{chan2022causalscrubbing}: When performing a forward pass we ablate half of the APD model's parameter components, excluding the ones that correspond to the currently active inputs (the `scrubbed' run). We compare this to ablating half of the parameter components \textit{including} those that correspond to currently active inputs (`anti-scrubbed'). Figure \ref{fig:weight-linearity} gives a visual illustration of the output of multiple `scrubbed' and `anti-scrubbed' runs for a one-hot input $x_{42} = 1$. We see that ablating unrelated components perturbs the output only slightly, barely affecting the overall shape.

\begin{figure}
    \centering
    \includegraphics[width=1\linewidth]{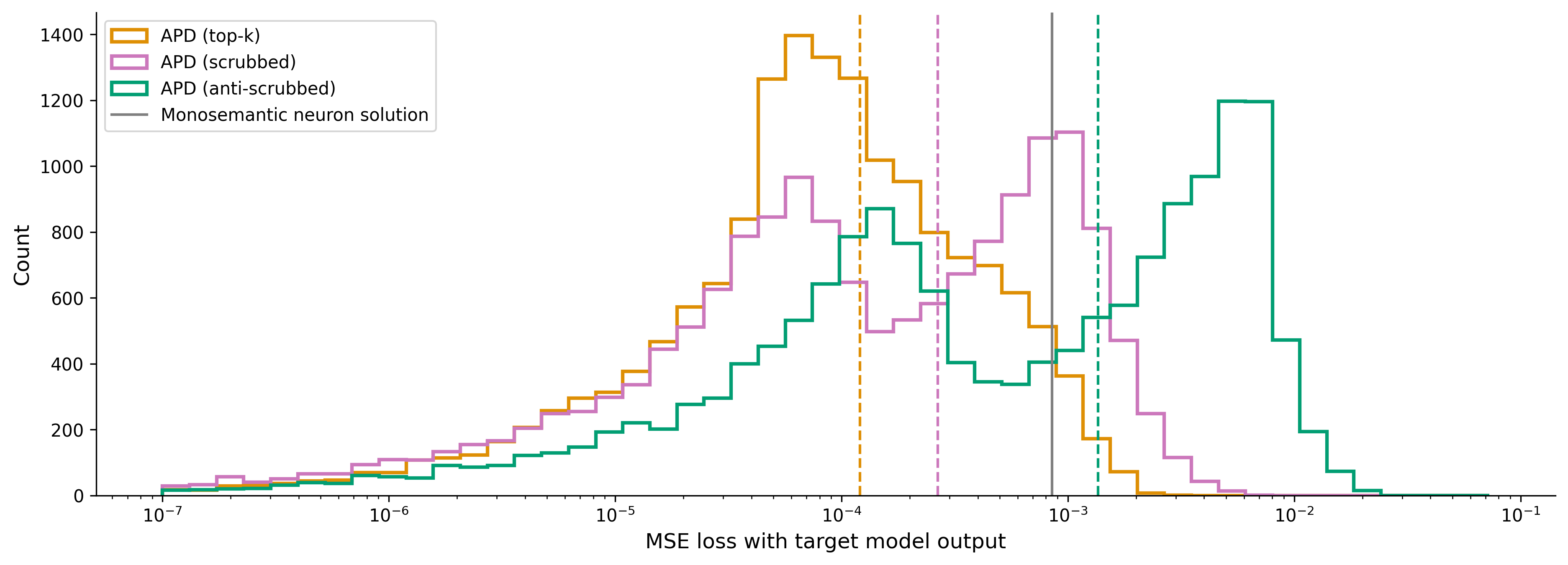}
    \caption{MSE losses of the APD model on the sparse forward pass (``top-$k$") and the APD model when ablating half ($50$) of its parameter components (``scrubbed" when none of the components responsible for the active inputs are ablated and ``anti-scrubbed" when they are ablated). The gray line indicates the loss for a model that uses one monosemantic neuron per input feature. The dashed colored lines are the mean MSE losses for each type of run.}
    \label{fig:scrubbing}
\end{figure}

To investigate whether this holds true for all components and on the training data distribution, we collect MSE losses into a histogram (Figure \ref{fig:scrubbing}). We find that the `scrubbed' runs (i.e. ablating unrelated parameter components -- pink histogram) does not cause a large increase in the MSE loss with respect to target network outputs. On the other hand, the anti-scrubbed runs (i.e. ablating parameter components that are deemed to be responsible for the computation -- green histogram) does cause a large increase in MSE. This suggests that parameter components have mostly specialized to implement the computations for particular input features.  

However, some parameter components appear to partially represent secondary input feature computations. This causes the visibly bimodal distributions of the scrubbed runs that can be seen in the figure: When these components are ablated, the loss of the model may be high when the secondary input feature is active. These components have the opposite effect on the loss when they are not ablated in the anti-scrubbed runs, making both scrubbed and anti-scrubbed losses bimodal. Preliminary work suggests that this can be improved with better hyperparameter settings or with adjustments to the training process, such as using alternative loss functions (Appendix \ref{app:l_p_loss}) or enforcing the APD components to be rank-$1$. We further quantify the extent to which some components partially represent secondary input feature computations in Appendix \ref{app:resid_mlp_1layer_apd}.

\subsection{Toy Model of Cross-Layer Distributed Representations}
\label{sec:residmlp_2layer}
We have seen how APD can learn parameter components that represent computations on individual input features, even when those computations involve neurons that contribute to the computations of multiple input features (Section \ref{sec:residmlp_1layer}). However, those computations take place in a single MLP layer. But realistic neural networks seem to exhibit cross-layer distributed representations \citep{yun2021sparse, lindsey2024crosscoders}. In this section, we show how APD naturally generalizes to learn parameter components that represent computations that are distributed across multiple MLP layers\footnote{For further validation that APD can identify cross-layer distributed representations, we apply it to a hand-coded network that implements a gated trigonometric function (Appendix \ref{app:handcoded_gated_model}).}.

\begin{figure}
    \centering
    \includegraphics[width=0.7\linewidth]{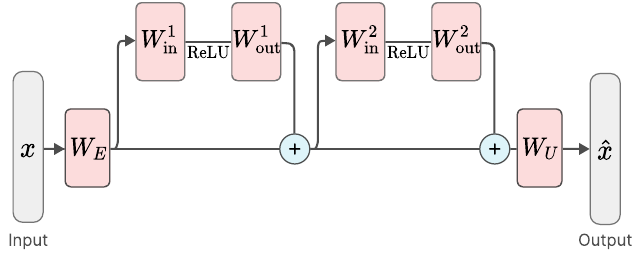}
    \caption{The architecture of our Toy model of Cross-Layer Distributed representations using a 2-layer residual MLP. We fix $W_E$ to be a randomly generated matrix with unit norm rows, and $W_U={W_E}^T$.}
    \label{fig:resid_mlp_2l_architecture}
\end{figure}

We extend the model and task used in the previous section by adding an additional residual MLP layer (Figure \ref{fig:resid_mlp_2l_architecture}). This model still performs compressed computation, but now with representations that are distributed across multiple layers. In this model, $W_E$ is again a fixed, randomly generated embedding matrix with unit norm rows and $W_U=W_E^T$. We keep the residual stream width of $d_{\rm resid}=1000$ and $100$ input features, but our $50$ MLP neurons are now split across layers, with $25$ in each of the two MLPs. We train APD on this model with $200$ parameter components (allowing for $100$ to die during training).

\subsubsection*{APD Results: Toy Model of Cross-Layer Distributed Representations}
\begin{figure}
    \centering
    \includegraphics[width=1\linewidth]{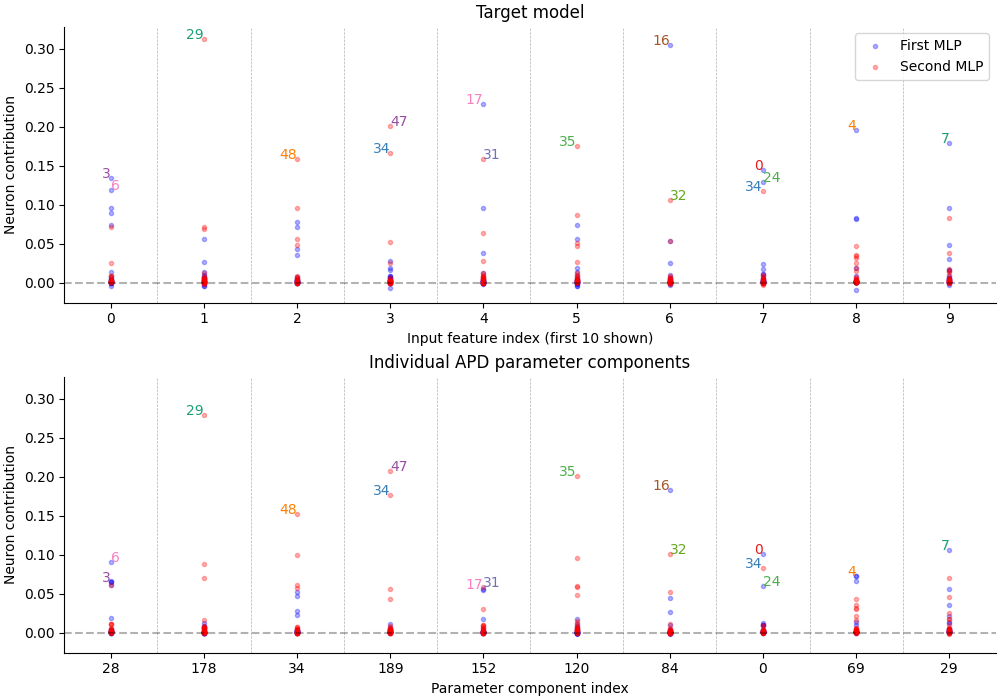}
    \caption{Similarity between target model weights and APD model components for the first $10$ input feature dimensions in a $2$-layer residual MLP. \textbf{Top}: Neuron contributions measured by $(W_E W_\text{IN}) \odot (W_\text{OUT} W_U)$ where $\odot$ is an element-wise product and $W_\text{IN}$ and $W_\text{OUT}$ are the MLP input and output matrices in each layer concatenated together.  \textbf{Bottom}: Neuron contributions for the learned parameter components, measured by $\max_k [({W_U}_{[i,:]} {W_\text{OUT}}_k) \odot ({W_\text{IN}}_k {W_E}_{[:,i]})]$ for each feature index $i\in[0,9]$. The neurons are numbered based on their raw position in the network, with neurons $0$ to $24$ in the first layer and neurons $25$ to $49$ in the second layer. An extended version of this figure showing all input features and parameter components can be found \href{https://api.wandb.ai/links/apollo-interp/h5ekyxm7}{here}.}
    \label{fig:resid-mlp-weights-2layers}
\end{figure}

APD finds qualitatively similar results to the $1$-layer toy model of compressed computation presented in Section \ref{sec:residmlp_1layer}. We see that the APD model learns parameter components that use neurons with large contribution values in both MLP layers (Figure \ref{fig:resid-mlp-weights-2layers}, bottom). Again, we find that the computations occurring in each parameter component closely correspond to individual input feature computations in the target model (Figure \ref{fig:resid-mlp-weights-2layers}, top versus bottom). For confirmation that the target model and APD model in the $2$-layer distributed computation setting yield results that closely matching those observed in the $1$-layer scenario, see Appendix \ref{app:resid_mlp_2layer_target} and Appendix \ref{app:resid_mlp_2layer_apd}, respectively. 

However, the results exhibit a larger number of imperfections compared to the $1$-layer case. In particular, more components represent two input feature computations rather than one. As in the $1$-layer case, we again notice that batch top-$k$ can cause some parameter components to not fully represent the computation of an input feature, and instead rely on activating multiple components for some input features (see Appendix \ref{app:resid_mlp_2layer_apd})\footnote{Further analysis can be found in the ``Toy Model of Cross-layer Distributed Representations ($2$ layers)'' section of this WandB \href{https://api.wandb.ai/links/apollo-interp/h5ekyxm7}{report}.}.

%% file: 5_Discussion.tex
\section{Discussion}\label{sec:discussion}

We propose APD, a method for directly decomposing neural network parameters into mechanistic components that are faithful, minimal, and simple. This takes a ‘\textit{parameters-first}’ approach to mechanistic interpretability. This contrasts with previous work that typically takes an ‘\textit{activations-first}’ approach, which decomposes networks into directions in activation space and then attempts to construct circuits (or ‘mechanisms’) using those directions as building blocks \citep{olah2020zoom, cammarata2020curve, cunningham2023sparse, bricken2023monosemanticity, marks2024sparse}.

A parameters-first approach has several benefits. The new lens it provides suggests straightforward solutions to many of the of the challenges presented by the activations-first approach to mechanistic interpretability. Nevertheless, it also brings novel challenges that will need to be overcome. In this section, we discuss both the potential solutions and challenges suggested by this new approach and suggest potential directions for future research.

\subsection{Addressing issues that seem challenging from an activations-first perspective}\label{sec:addressing_activation_challenges}

\paragraph{The activations-first paradigm struggles to identify minimal circuits in superposition, while APD achieves this directly.} 

APD optimizes a set of parameter components that are maximally simple while requiring as few as possible to explain the output activations of any given datapoint. It is possible to think about these parameter components as circuits, since they describe transformations between activation spaces that perform specific functional roles. 

Identifying a method to obtain minimal circuits by building on sparse dictionary learning (SDL) -- which is an activations-first approach -- has proven difficult for several reasons. One reason is that even though SDL might identify sparsely activating latent directions, there is no reason to expect the connections between them to be sparse. This might result in dense interactions between latents in consecutive layers, which may be difficult to understand compared with latent directions that were optimized to interact sparsely. Another reason that SDL has struggled to identify concise descriptions of neural network parameters is the phenomenon of feature splitting \citep{bricken2023monosemanticity}, where it is possible to identify an ever larger number of latents using ever larger sparse dictionaries. However, more latents means more connections between them, even if the transformation implemented by this layer is very simple! Descriptions of the connections may include an ever growing amount of redundant information. As a result, even very simple layers that transform latents in superposition may require very long description lengths. 

\paragraph{A parameters-first approach suggests a conceptual foundation for ‘features’.}

A central object in the activations-first paradigm of mechanistic interpretability is a ‘feature’. Despite being a central object, a precise definition remains elusive. Definitions that have been considered include \citep{elhage2022toy}: 
\begin{enumerate}
    \item \textit{‘Features as arbitrary functions’}, but this fails to distinguish between features that appear to be fundamental abstractions (e.g. a ‘cat feature’) and those that don’t (e.g. a ‘cat OR car’ feature). 
    \item \textit{‘Features as interpretable properties’}, but this precludes features for concepts that humans don’t yet understand.
    \item \textit{‘Features as properties of the input which a sufficiently large network will reliably dedicate a neuron to representing’}. This definition is somewhat circular, since it defines object in neural networks using objects in other neural networks, and may not account for multidimensional features \citep{engels2024languagemodelfeatureslinear, olah2024linearmultidimensional} . 
\end{enumerate}

In our work, we decompose neural networks into parameter components that minimize mechanistic description length, which we call the network's `mechanisms'. Note that a network's mechanisms are not equivalent to its `features', but they might be related. Defining a network's features as `\textit{properties of the input that activate particular mechanisms}' 
seems to overcome the definitional issues above.

In particular, it overcomes the issues in Definition 1 because a set of maximally faithful, minimal, and simple components should learn to correspond to ‘cat mechanisms’ and ‘car mechanisms’, but not ‘cat OR car mechanisms’ (unless the target network actually did have specific machinery dedicated to ‘cat OR car’ computations). APD also does not rely on a notion of human interpretability, overcoming the issue with Definition 2. It also seems to overcome the issues of Definition 3, since the definition is not circular and should also be able to identify multidimensional mechanisms (and hence multidimensional features that activate them), although we leave this for future work.

The definition also overcomes issues caused by `feature splitting', a phenomenon observed in SDL where larger dictionaries identify sets of different features depending on dictionary size, with larger dictionaries finding more sparsely activating, finer-grained features than smaller dictionaries. This happens because SDL methods can freely add more features to the dictionary to increase sparsity even if those features were not fundamental building blocks of computation used by the original network. APD components also need to be faithful to the target network's parameters when they are added together, meaning it cannot simply add more components in order to increase component activation sparsity or simplicity. To see this, consider a neural network that has a hidden layer that implements a $d$-dimensional linear map on language model data. A transcoder could learn ever more sparsely activating, ever more fine-grained latents to minimize its reconstruction and sparsity losses and represent this transformation. By contrast, the APD losses would be minimized by learning a single $d$-dimensional component that performs the linear map. The losses cannot be further reduced by adding more components, because that would prevent the components from summing up to the original network weights. 

Incidentally, this thought experiment not only sheds light on feature splitting, but also sheds light on the difference between parameter components and `features' as they are usually conceived. Parameter components are better thought of as ``steps in the neural network's algorithm'', rather than ``representations of properties of the input''. The network may nevertheless have learned mechanisms that specifically activate for particular properties of the input, which may be called `features'.

\paragraph{A parameters-first approach suggests an approach to better understanding ‘feature geometry’.}
\cite{bussman2024metasaes} showed that the Einstein SAE latent has a similar direction to other SAE latents that were German-related, physics-related, and famous people-related. This suggests that the latents that SDL identify lie on an underlying semantic manifold. Understanding what gives this manifold its structure should suggest more concise descriptions of neural networks. But SDL treats SDL latents as fundamental computational units that can be studied in isolation, thus ignoring this underlying semantic manifold \citep{Mendel_2024}. 
We contend that the reason the Einstein latent points in the ‘physics direction’ (along with other physics-related latents) is because the network applies ‘physics-related mechanisms’ to activations along that direction. Therefore, by decomposing parameter space directly, we expect interpretability in parameter space to shed light on computational structure that gives rise to a network's SAE `feature geometry'.

\paragraph{Interpretability in parameter space suggests an architecture-agnostic method to resolving superposition.}
Neural network representations appear not to neatly map to individual architectural components such as individual neurons, attention heads, or layers. Representations often appear to be spread across various architectural components, as in attention head superposition \citep{jermyn2023attentionheadsuperposition} or cross layer distributed representations \citep{yun2021sparse, lindsey2024crosscoders}. It is unclear how best to adapt SDL to each of these settings in order to tell concise mechanistic stories \citep{mathwin2024gated, wynroe2024qkbilinear, lindsey2024crosscoders}. A more general approach that requires no adaptation would be preferred. Interpretability in parameter space suggests a way to overcome this problem in general, since any architecture can in theory be decomposed into directions in parameter space without the method requiring adaptation.

\subsection{Next steps: Where our work fits in an overall interpretability research and safety agenda}\label{sec:discussion_agenda}

We had two main goals for this work. Our first goal was to resolve conceptual confusions arising in the activations-first, sparse dictionary learning-based paradigm of mechanistic interpretability.
Our other main goal was to develop a method that builds on these conceptual foundations that can be applied to real-world models. However, APD is currently only appropriate for studying toy models because of its computational cost and hyperparameter sensitivity. We see two main paths forward:
\begin{enumerate}
    \item \textbf{Path 1:} Develop APD-like methods that are more robust and scalable. 
    \item \textbf{Path 2:} Use the principles behind our approach to design more intrinsically decomposable architectures\footnote{In particular, we are excited about research that explores how to pre-decompose models using mixtures-of-experts with many experts, where the experts may span multiple layers, like the parameter components in our work.}. 
\end{enumerate}

We are excited about pursuing both of these paths. In the rest of this section, we focus on Path 1, leaving Path 2 to future work.

We will outline what we see as the main challenges and exciting future research directions for building improved methods for identifying minimal mechanistic descriptions of neural networks in parameter space. To become more practical, APD must be improved in several ways. In particular, it should have a lower computational cost; be less sensitive to hyperparameters; and more accurate attributions. We may also need to fix outlying conceptual issues, such as the extent to which APD privileges layers.  We also discuss several safety-oriented and scientific research directions that we think may become easier when taking a parameter-first approach to interpretability.

\subsubsection{Improving computational cost} \label{sec:computational_cost_improvement}

While developing this initial version of APD, we focused on conceptual progress over computational efficiency. 
At a glance, our method involves decomposing neural networks into parameter components that each have a memory cost similar to the target network. That would make it very expensive, scaling in the very worst case as something like $\mathcal{O}(N^2)$ where $N$ is the parameter count of the original model.
However, there are several reasons to think this might not be as large an issue as it initially appears:
\begin{itemize}
    \item \textbf{More efficient versions of APD are likely possible.} We think there exist paths toward versions of APD that are more computationally efficient.
    For the experiments in this paper, we often permitted the parameter components to be full rank. But theories of computation in superposition suggest that for a network to have many non-interfering components, they need to be low rank or localized to a small number of layers \citep{Bushnaq_Mendel_2024}. A version of APD that constrains components to be lower rank and located in fewer layers would reduce their memory cost. Even if there are many high rank mechanisms, we think it may be possible to identify principles that let us stitch together many low rank components if the ground truth mechanisms are high rank, or let us use hierarchical representations of components. It may be possible to apply our method to models one layer at a time, like transcoders, which may save on memory costs of having to decompose every parameter at the same time (discussed further in Section \ref{sec:relatedwork_transcoders}).
    \item \textbf{Alternative approaches, such as SDL, may be even more expensive. } To use sparse dictionary learning to decompose a single layer’s activation space may be relatively cheap compared with training an entire network. But even if it were possible to reverse engineer neural networks using sparse dictionaries (which is unclear), we would need to train sparse dictionaries on every layer in a network in order to reverse engineer it, which may be very expensive. It becomes even more expensive considering the need to identify or learn the connections between sparse dictionary latents in subsequent layers. At present, there is no reason to expect that it costs less to train sparse dictionaries on every layer than to perform APD. It may indeed cost much more to use SDL, since we do not know in advance what size of dictionary we need to use and how much feature splitting to permit. We suspect that a reasonably efficient version of APD, which aims to identify minimal mechanistic descriptions, to reverse engineer networks will fare favorably compared to using SDL to achieve similar feats, if SDL can be used for that purpose at all.
    \item \textbf{Our method suggests clear paths to achieving interpretability goals that other approaches have struggled to achieve.} Even if our method were more expensive than SDL-based approaches, our approach confers significant advantages (discussed in Section \ref{sec:addressing_activation_challenges}) that might make the computational cost worth it.
\end{itemize}

\subsubsection{Improving robustness to hyperparameters}
A practical issue at present is that the method is sensitive to hyperparameters. Extensive hyperparameter tuning was often required for APD to find the correct solution. Making the method more robust to hyperparameters is a high priority for future work. It is worth noting that we encountered fewer hyperparameter sensitivity issues when scaling up the method to larger toy models. It is likely that hyperparameter sensitivity was exacerbated due to the amount of interference noise between input feature computations in our experiments in small dimensions, and this may resolve itself when scaling up.

\subsubsection{Improving attributions}
One of the reasons that the method might not be robust is that the method currently uses gradient attributions, which are only a first order approximation of causal ablations \citep{mozer1988skeletonization, molchanov2017pruningconvolutionalneuralnetworks}. Previous work, such as AtP \citep{nanda2022attribution} and AtP* \citep{kramár2024atpefficientscalablemethod} indicates that using gradients as first-order approximations to causal ablations work reasonably well, but become unreliable when gradients with respect to parameters become small due to e.g. a saturated softmax \citep{kramár2024atpefficientscalablemethod}. This problem could potentially be alleviated by using integrated gradients \citep{sundararajan2017axiomatic}, learning masks for each parameter component on each datapoint during training \citep{caples2024scalingsparse}, or other attribution methods instead.

\subsubsection{Improving layer non-privileging}\label{app:non-privileging}
We want to find components in the structure of the learned network algorithm, rather than the network architecture. 
Thus, we would prefer our formalism to be entirely indifferent to changes of network parameterization that do not affect the underlying algorithm. 
However, APD is currently not indifferent to changes of network parameterization that mix network layers. Layers are therefore still slightly privileged by APD.
This is because we optimize parameter components to be simple by penalizing them for being high rank, and the rank of weight matrices cannot be defined without reference to the network layers. 
Thus, if two components in different neural networks implement essentially the same computation, one in a single layer, the other in cross-layer superposition, the latter component may be assigned a higher rank. Therefore, while we think that APD can still find components that stretch over many layers, it may struggle to do so more than for components that stretch over fewer layers. 
We would need to find layer-invariant quantities that more accurately track the simplicity of components independent of their parametrization than effective rank. Speculatively, some variation of the weight-refined local learning coefficient \citep{wang2024differentiationspecializationattentionheads} might fulfill this requirement.

\subsubsection{Promising applications of interpretability in parameter space}

If we can overcome these practical hurdles, we think that interpretability in parameter space may make achieving some of the safety goals of interpretability easier than with activations-first methods. For instance, if we have indeed identified a way to decompose neural networks into their underlying mechanisms, it will be readily possible to investigate mechanistic anomaly detection for monitoring purposes \citep{christiano2022mad}. Interpretability in parameter space may also be easier to perform precise model editing or unlearning of particular mechanisms [e.g. \cite{meng2023locating, meng2023masseditingmemorytransformer}], since model descriptions are given in terms of parameter vectors, which are the objects that we would directly modify. 

We are also excited about applications of APD that might help answer important scientific questions. For instance, we suspect that APD can shed light on the mechanisms of memorizing vs. generalizing models \citep{Henighan2023memorization, zhang2017understandingdeeplearningrequires, arpit2017memorization}; the mechanisms of noise robustness \citep{morcos2018importance}; or leveraging the fact that APD is architecture agnostic in order to explore potentially universal of mechanistic structures that are learned independent of architecture \citep{li2015convergent, olah2020zoom}, such as convolutional networks, transformers, state space models, and more.

\section{Conclusion}\label{sec:conclusion}

This work introduces Attribution-based Parameter Decomposition (APD) as a fundamental shift in mechanistic interpretability: instead of analysing neural networks through their activations, we demonstrate that directly decomposing parameter space can reveal interpretable mechanisms that are faithful, minimal, and simple. Our approach suggests solutions to long-standing problems in mechanistic interpretability, including identifying minimal circuits in superposition, providing a conceptual foundation for `features', enabling better understanding of `feature geometry', and serving as an architecture-agnostic approach to neural network decomposition.

Our experiments demonstrate that APD can successfully identify ground truth mechanisms in multiple toy models: recovering features from superposition, separating compressed computations, and discovering cross-layer distributed representations.

Although our results are encouraging, several challenges remain before APD can be applied to real-world models. These include improving computational efficiency, increasing robustness to hyperparameters, and incorporating more robust attribution methods. 

By decomposing neural networks into their constituent mechanisms, this work brings us closer to reverse engineering increasingly capable AI systems, helping to open the door toward a variety of exciting scientific and safety-oriented applications.

\section{Related Work}\label{sec:related_work}

Our approach draws on several ideas from prior work in mechanistic interpretability and other fields. 

\subsection{Sparse Autoencoders}
Sparse Autoencoders (SAEs) are a sparse dictionary learning (SDL) method that is commonly used in mechanistic interpretability. Although APD is not a SDL method, it has many connections. 

SAEs can be used to identify an overcomplete basis for activation space consisting of sparsely activating directions  \citep{lee2007sparse,yun2021sparse,Sharkey_Braun_Millidge_2022,cunningham2023sparse,bricken2023monosemanticity}. Similarly, APD finds a set of sparsely used vectors (parameter components) in parameter space. However, we do not expect this set of vectors to form an overcomplete basis for parameter space, since the number of mechanisms a network can implement is upper-bounded by its parameter count \citep{Bushnaq_Mendel_2024}.

Networks that have sparsely activating latent directions might sometimes dedicate a mechanism to operate on one of these directions. See Section \ref{sec:TMS_results} for an example. 
However, parameter components are more general than directions in activation space. They may also operate on multidimensional activation subspaces, stretch over multiple layers, and operate on input latents that are not sparse.

The training process used in our work, notably the second forward pass that uses only the (batch) top-$k$ most attributed parameter components, resembles the training process of top-$k$ and batch top-$k$ sparse autoencoders \citep{makhzani2013k, gao2024scalingevaluatingsparseautoencoders, bussmann2024batchtopk}. An alternative APD procedure that uses an $L_1$ or $L_p$ penalty on attributions is also possible (Appendix \ref{app:l_p_loss}), which would be analogous to classic SAE training methods with an $L_1$ or $L_p$ penalty on latent activations. 

Crosscoders are a variant of SAEs that take as input and reconstruct activations at multiple layers simultaneously. They can identify representations that span multiple layers \citep{lindsey2024crosscoders, yun2021sparse}. Similarly, our work identifies mechanisms that span multiple layers. However, like other SDL approaches, crosscoders decompose the activations, which are the \textit{result} of a network's mechanisms, and do not immediately suggest a way to decompose the network's mechanisms themselves. 

\cite{braun2024identifying} train end-to-end SAEs to identify latent directions for which as few as possible are necessary to reconstruct the output activations and hidden activations. APD similarly identifies parameter components for which as few as possible are necessary to reconstruct the output activations (and sometimes the hidden activations).

SAEs can be considered as lossy compression algorithms that optimize for compressed descriptions of activation datasets \citep{ayonrinde2024interpretabilitycompressionreconsideringsae}. Similarly, APD optimizes for compressed descriptions of the causal processes that a network applies to activations over the course of computing its output. A somewhat related perspective on neural network decomposition and interpretation is proposed in \cite{gross2024compactproofsmodelperformance}, which argues that the amount of mechanistic understanding about a neural network can be meaningfully quantified by how much it compresses proofs about the network's behavior.

\subsection{Transcoders}\label{sec:relatedwork_transcoders}
Transcoders are an SDL method that is similar to SAEs but is trained to reconstruct the output activations of a layer given its input activations \citep{dunefsky2024transcodersinterpretablellmfeature, mathwin2024gated, wynroe2024qkbilinear}.
Although transcoders decompose activations, they can also be considered to decompose the transformation implemented by that layer. This makes them related to APD, though APD decomposes the transformation in parameter space.

However, the `activation' of parameter components are disanalogous to the activation of transcoder latents. In a transcoder, the activation of a latent is determined by whether an activation on the forward pass is above a threshold. This nonlinear threshold filters out the `interference terms' from non-orthogonal latents represented in superposition and prevents them from being forward-propagated. In APD, the `activation’ of a parameter component is determined by whether it had a causal influence downstream (i.e. whether it was attributed). In APD, attributions, rather than nonlinearities, filter the interference terms from non-orthogonal parameter components and prevent their effects from being forward-propagated. 

Although our work decomposed all layers of our toy models at once, it is likely possible to apply APD to only one layer at a time. This approach may be useful for keeping computational costs low (see Section \ref{sec:computational_cost_improvement} for further discussion). Similarly, transcoders could in theory be trained on every layer at the same time, though this may be even more expensive than APD, especially since there is no upper limit on the number of latents that transcoders should have.

With any finite number of dictionary elements, transcoders will always involve an irreducible activation reconstruction error term, even in the limit of using infinite latents \citep{engels2024decomposingdarkmattersparse}. This is because transcoders attempt to use linear combinations of activation directions to approximate the transformations implemented by a layer, such as an MLP layer \citep{dunefsky2024transcodersinterpretablellmfeature} or attention block \citep{mathwin2024gated, wynroe2024qkbilinear} (among other reasons). But those may use very different activation functions, such as a softmax nonlinearity or GeLUs and may be difficult to approximate with any finite number of dictionary elements. 
APD therefore has the additional benefit of being equally applicable to arbitrary neural architectures, such as attention blocks, recurrent neural networks, convolutional networks, or state space models, without requiring specific adaptation to each architecture.

\subsection{Weight masking and pruning}
Some previous work identifies masks for neural network parameters in order to isolate particular functionality. This is similar to APD, since activating a parameter component is functionally equivalent to leaving that component unmasked while masking others. Our approach is therefore similar to learning many different parameter masks, one for each parameter component. 

\cite{mozer1988skeletonization} is an early work that identifies relevant units in a neural network and prunes irrelevant ones, thus implicitly masking the parameters that connect to those units. Similar to our work, it uses gradient-based attributions to identify relevant components. Similarly, \cite{lecun1989optimalbraindamage} ablates individual parameters, but uses second-order approximations of causal perturbation in contrast to our first-order approximations. Later work identifies explicit or implicit binary masks (or masks whose elements take values in $[0, 1]$) over units or parameters \citep{csordás2021neuralnetsmodularinspecting, cao2021lowcomplexityprobingfindingsubnetworks, zhang2021subnetworkstructurekeyoutofdistribution, decao2021sparseinterventionslanguagemodels, patil2023neuralsculpting, lepori2023breakdownevidencestructural, mondorf2024circuitcompositionsexploringmodular}. However, masks that are constrained to take values in $[0,1]$ privilege components aligned with the neuron- or parameter-bases. But gradient descent does not necessarily privilege these bases. Our implicit masks may take values outside $[0,1]$ so that they do not privilege components that align with the neuron- or parameter-bases\footnote{It may be somewhat counterintuitive to suppose that we can decompose network parameters into components that take positive or negative values when a given parameter is e.g. positive. But other parameter decomposition methods (such as singular value decompositions of matrices) also permit this while nevertheless being meaningful functional decompositions.}. 

\cite{csordás2021neuralnetsmodularinspecting} learned differentiable binary masks over weights to identify components responsible for specific functions. However, this required collecting a dataset to define a task distribution that defines which components would be used. This is unlike our work, which optimizes components for minimum description length on any given datapoint, which implicitly defines task distributions on which particular components are used in an unsupervised way.  

\subsection{Circuit discovery and causal mediation analysis}

Circuit discovery methods are a broad class of approaches that typically use causal interventions \citep{chan2022causalscrubbing, wang2022interpretability, conmy2024towards} (or an approximation of them \citep{nanda2022attribution, syed2023attributionpatchingoutperformsautomated, kramár2024atpefficientscalablemethod}) on activations to find simple circuits that are sufficient for computations required for performance on particular tasks if the remainder of the model is ablated. 

These approaches and APD have many similarities, but also crucial differences. They both involve ablating components of models with the aim of finding core computational mechanisms. However, previous circuit dicsovery methods have typically assumed a particular decomposition of networks into neurons, MLPs, attention heads, while APD makes much fewer assumptions about components. Later work uses sparse dictionary learning to determine the components that are causally intervened upon \citep{cunningham2023sparse, bricken2023monosemanticity, marks2024sparse, geiger2024findingalignmentsinterpretablecausal}. However, these approaches assumed components in activation space that are localized in single layers. APD learns components in parameter space. 

Circuit discovery methods operate on a computational graph, representing all components of a network and their interactions. But the way in which components interact is abstracted away; typically every connection is simply assigned a scalar interaction strength. APD on the other hand operates on the parameters, with each parameter component containing all the parameters needed to explain its function.  
Circuit discovery methods also measure complexity as the number of nodes or edges in the computational graph, while APD uses an approximation of the sum of the rank of the weight matrices in the parameter component.

\subsection{Interpreting parameters}
Although most prior work aims to understand activations primarily, there exists a small amount of work that tackles weights more directly. 

Some early machine learning \citep{rumelhart1986bp, mcclelland1985distributed, srivastava2014dropout} and mechanistic interpretability \citep{olah2020overview} interprets the raw parameters, thus assuming a decomposition that aligns with the unit basis in parameter space. 

Other work decomposes the parameters of the network using matrix decomposition methods. \cite{millidge2022svd}, \cite{meller2023svr} and \cite{gross2024compactproofsmodelperformance} use SVD to decompose parameters. Although these works found interpretable structure, SVD is not capable of identifying mechanisms in superposition and is limited to single layers, while mechanisms may span multiple layers. \cite{voss2021visualizing} and \cite{petrov2021weight} both use NMF to decompose weights of image classifier models into factors and visualize the results. \cite{voss2021visualizing} go further and study expanded weights, which are the result of multiplying weights of adjacent layers which, in practice, uses gradients. However, the factorization approaches used in this work do not decompose parameters into components that are minimal and simple, even if they may be faithful (see also Section \ref{sec:TMS_results} for further comparison between APD and NMF).

Static analysis of parameters studies the computational structure of a network without needing to run forward or backward passes. One example is \cite{dar2023analyzingtransformersembeddingspace}, who use the logit lens \citep{nostalgebraist2020logit} to project model parameters into vocabulary space to help interpret them. Other work interprets decompositions of the weights of bilinear layers, since their simple mathematical form facilitates decomposition despite their nonlinearity \citep{pearce2024bilinearmlpsenableweightbased, pearce2024weightbaseddecompositioncasebilinear}.

\subsection{Quanta identification}
\cite{michaud2024quantizationmodelneuralscaling} identify `quanta', where a quantum is defined as `An indivisible computational module that, for example, retrieves a fact, implements an algorithm, or more generally corresponds to some basic skill possessed by a model'. This definition reflects aspects of what our work defines as a `mechanism', though our definition is formal. To identify quanta in language models, they cluster the gradients of the loss with respect to the parameters. This shares the intuition with our work that different gradients reflect different active mechanisms. However, clustering gradients does not decompose them. By training parameter components to be sparsely attributed, our method implicitly trains parts of our parameter components to align sparsely with components of gradients (Section \ref{sec:TMS_results}).

\subsection{Mixture of experts}
Mixture of expert (MoE) architectures leverage the notion that only a subset of network's mechanisms are useful on any given forward pass \citep{jacobs1991mixtureofexperts}. Each expert is used only on a subset of the training dataset, and tends to specialize to that subdistribution, similar to the mechanisms learned in our method \citep{fedus2022reviewsparseexpertmodels}. However, MoE architectures usually also have a gating function that determines which experts are used during the forward pass. Our approach, by contrast, can only identify which experts are active using attribution methods, which require an initial forward pass. Another difference is that our mechanisms may span multiple layers, whereas experts in MoE architectures typically exist in one layer only (though see \cite{park2024monetmixturemonosemanticexperts}). The principles uncovered by our method may be useful to develop new mixture-of-experts approaches that are more easily decomposed and interpreted.

\subsection{Loss landscape dimensionality and degeneracy}
Our work is directly inspired by research on the intrinsic dimensionality of and degeneracy in the loss landscape. Both our work and \cite{li2018measuring} parametrize a network using linear combinations of basis vectors in parameter space. However, \cite{li2018measuring} used a fixed, random basis for a subspace of parameter space and trained only the coefficients. Our work, by contrast, trained the bases such that they are simple and uses as few basis elements as possible for any given datapoint. 

Singular learning theory \citep{watanabe2009algebraic} quantifies degeneracy in network weights present over the entire data distribution using the learning coefficient. APD finds vectors in parameter space along which parameters can be ablated on some network inputs, so they are degenerate directions over a subset of the data. Likewise, \cite{wang2024differentiationspecializationattentionheads} defined the data-refined learning coefficient, which measures degeneracy in neural network weights over a subset of the distribution. Unlike \cite{wang2024differentiationspecializationattentionheads}, APD finds subsets of the data over which some directions are degenerate in an unsupervised manner rather than starting with a subset of the distribution and quantifying its degeneracy.

\section*{Author contributions}\label{sec:author-contributions}

\paragraph{Research iteration}
Our method underwent significant iteration throughout development, changing many times in response to experimental results. DB, SH, LB, JM, and LS were responsible for driving forward various steps in the iteration cycle that led to the current paper.

\paragraph{Conceptualisation}
The core ideas for APD were developed through close collaboration between LS and LB.
LS originated the idea to decompose the network parameters into sparsely used directions in parameter space and made initial suggestions for the faithfulness, minimality, and simplicity losses. SH and JM red-teamed the idea that directions in parameter space correspond to individual computations in networks with polysemanticity or computation in superposition.
LB, with input from JM, developed a better minimality loss based on attributions and developed the MDL framing and its formalism.
LS developed the idea for the top-k version of the method (versus the original $L_p$ penalty version).
The idea for the Schatten-norm-based simplicity loss was developed independently by LB and LS, based on experiments from DB and SH. LB developed the idea for an efficient implementation of that loss.
SH, JM, LS, and LB identified various problems with previous versions of the method during the research iteration cycle.

\paragraph{Target models and task designs}
LB and LS independently identified the Toy Model of Superposition as a good toy model for APD. 
LB, SH, and DB developed the Toy Model of Compressed Computation, with implementation and training by DB and SH.
JM and LB developed the idea for the handcoded model in the appendix, and JM handcoded it.

\paragraph{Experiments and analysis}
DB ran most of the experiments throughout the various research iteration cycles. DB and SH ran most of the experiments for Toy Model of Compressed Computation and Toy Model of Cross Layer Distributed Representations. DB and SH analyzed most of the experiments, with significant input from LB and some input from LS. LB helped analyse experiments through the research iteration cycle. 
SH and JM ran most of the experiments for the hardcoded model throughout the iteration cycle; DB and SH ran the experiments on the version that is in the appendix. 

\paragraph{Writing}
LS, LB, DB, and SH contributed to writing and editing the paper. LS wrote most of the Introduction, Discussion, Conclusion, and Related Work. LS and LB wrote the methods section. LB wrote most of the more detailed description of the APD method in the appendix. DB wrote most of the results section and appendix results, with significant contributions from SH and LS.

\paragraph{Figures and illustrations}
LS designed and made the illustrations of the method and target models’ architectures. SH and DB designed and generated the results figures.

\section*{Acknowledgements}

Our work benefited from valuable discussions with and feedback from many colleagues and collaborators.
We owe particular gratitude to Bilal Chughtai, Nicholas Goldowsky-Dill, Kaarel Hänni, and James Fox. We also owe thanks for their helpful feedback to Adrià Garriga-Alonso, Garret Baker, Joshua Batson, Brianna Chrisman, Jason Gross, Gurkenglas, Leo Gao, Marius Hobbhahn,
Linda Linsefors, Daniel Murfet, Neel Nanda, Michael Pearce, Dmitry Vaintrob, John Wentworth, Jeffrey Wu, and likely many others who we are regrettably forgetting.

%% file: 6_appendix.tex
\section{More detailed description of the APD method} \label{app:full_spd_explan}
Suppose we have a trained neural network $f(x,\theta^*)$, mapping network inputs $x$ to network outputs $y=f(x,\theta^*)$, with parameters $\theta^*\in \mathbb{R}^N$. 

We want to decompose into the `mechanisms' that the network uses to compute its behavior. The network's parameters implement these mechanisms. We would therefore like some way to decompose a network's parameters into its constituent mechanisms. 

We first define a set of parameter components 
\begin{equation}\label{eq:components}
    P=\{P_1,\dots,P_C\}, \quad P_c \in \mathbb{R}^{N}, \quad \forall c \in \text{range}(1,\dots,C), \quad C\in \mathbb{N}.
\end{equation}
We want to train these parameter components such that they correspond to a network's mechanisms. We think that it is reasonable to define a network's mechanisms as a set of components that minimizes the total description length of the network's behavior, per data point, over the training dataset. In particular, we want to identify components that are maximally faithful to the target network, maximally simple, and where as few as possible are used to replicate the network's behavior on any given datapoint.

In Section \ref{app:LPD}, we will more carefully define what we mean by a parameter vector $P_c$ being ‘used’ to
replicate the network’s behavior, and how to approximately measure this in practice using attribution
techniques. In Section \ref{app:MDL}, we will first discuss in what sense we want the average description length per datapoint to be minimised by deriving an idealized loss for APD. Then, we will use some approximations to find a proxy for this idealized loss that is more tractable to optimise in practice.

\tocless\subsection{Linear Parameter Decomposition}\label{app:LPD}
We want the components to be faithful to the original model in the sense that ideally, they should sum to form the target model's parameters:
\begin{equation}\label{eq:param_decomp}
    \theta^*=\sum^C_{c=1} P_c\,.
\end{equation}
However, we also want most of the parameter components $P_c\in \mathbb{R}^N$ to not be `used' on any one network input $x$, in the sense that we can ablate all but a few of them without  changing the outputs of the network. 

\paragraph{A definition of `inactive' parameter components}
We could try to operationalize the idea of most components being `inactive' in the sense of playing no meaningful role in the computation of the output by initializing a new network with a parameter vector $\kappa(x)$ composed of only a few of the most `used' parameter components, and demanding that:
\begin{equation}
\begin{aligned}
&f(x\vert \kappa(x))\approx f(x\vert \theta^*)
\end{aligned}
\end{equation}
where
\begin{equation}
\begin{aligned}
\kappa(x):=\sum^C_{c=1} s_c(x) P_c, \quad s_c(x) \in \{0,1\}, \quad \sum^C_{c=1} s_c(x)  \ll C.\\
\end{aligned}
\end{equation}

\paragraph{A stricter definition of `inactive' parameter components}
If the `inactive' components are not playing any meaningful role in the computation of the output, we should also be able to ablate or partially ablate them in any combination, and still get the same result. 
In general, we should get the same network output for any parameter configuration along any monotonic `ablation curve' $\gamma_c(x,t)$ where $t \in [0, 1]$:
\begin{equation}\label{eq:ablation_curves}
\begin{aligned}
\gamma_c(x,0)= 1, \quad &\gamma_c(x,1) = s_c(x), \quad \kappa(x,t):=\sum^C_{c=1} \gamma_c(x,t) P_c \\
&f(x\vert \kappa(x,t))\approx f(x\vert\theta^*)\,.
\end{aligned}
\end{equation}
This is a stricter definition of `inactive' that seeks to exclude cases like components $P_c$ and $P_{c'}$ both being `active' and important but canceling each other out. Without this stricter condition, we could have pathological solutions to the optimization problem. For example, suppose $P_1,\dots,\,P_{C-1}$ are a large set of parameter vectors for expert networks specialized to particular tasks that a target network $\theta^*$ is capable of. However, also suppose that these parameter vectors are completely unrelated to the underlying mechanisms in the target network $\theta^*$. Then, we could set the last component to
\begin{equation*}
\begin{aligned}
P_{C}=\theta^*-\sum^{C-1}_{c=1} P_{c}\,.
\end{aligned}
\end{equation*}
The resulting components would add up to the target network, $\sum_c P_c=\theta^*$. And a single $P_c$ would always be sufficient to get the same performance as the target network. However, the components $P_c$ could be completely unrelated to the mechanistic structure of the target network. Requiring that the parameter components can be ablated part of the way and, in any combination, excludes counterexamples like this.

Equations \ref{eq:param_decomp} and \ref{eq:ablation_curves} together define what we mean when we say that we want to decompose a network's parameter vector into a sum of other parameter vectors that correspond to distinct mechanisms of the network. The idea expressed in this definition is that different mechanisms combine \emph{linearly} in parameter space to form the whole network. We call the class of methods that attempt to decompose neural network parameters into components that approximately satisfy Equations \ref{eq:param_decomp} and \ref{eq:ablation_curves} \textit{Linear Parameter Decomposition} (LPD) methods. 

\paragraph{Component attributions}\label{app:attributions}

Directly checking that Equation \ref{eq:ablation_curves} is satisfied would be computationally intractable. Instead, for APD, we try to estimate whether the condition is approximately satisfied by calculating attributions of the output to each component. Currently, we do this using gradient attributions  \citep{finlayson2021causalanalysissyntacticagreement, molchanov2017pruningconvolutionalneuralnetworks, neel2022attribution}. This estimates the effect of ablating $P_c$ as:
\begin{equation}
A_c(x):= \sqrt{\frac{1}{d_L} \sum^{d_L}_{o=1} {\left(\sum_{l,i,j}\frac{\partial f_o(x,\theta^*)}{\partial \theta_{l,i,j}}P_{c,l,i,j}\right)}^2}\,.
\label{eq:attribution}
\end{equation}
We take the average square of this term over all output indices $o$, where the final output layer has width $d_L$.

Previous work, such as \citep{nanda2022attribution} and \citep{kramár2024atpefficientscalablemethod}, indicates that gradient-based first-order attribution methods can be somewhat accurate in many circumstances, but not always. For example, a saturated softmax in an attention head would render them inaccurate. Therefore, we expect that we might need to move to more sophisticated attribution methods in the future, such as integrated gradients \citep{sundararajan2017axiomatic}.

\paragraph{The assumption of parameter linearity} 
If neural networks do consist of a unique set of mechanisms in a meaningful sense, the ability of APD and any other LPD methods to recover that set of mechanisms relies on the assumption that the mechanisms are encoded in the network parameters in the linear manner specified by equations \ref{eq:param_decomp} and \ref{eq:ablation_curves}, at least up to some approximation. We call this the \textit{assumption of parameter linearity}.

The assumption of parameter linearity approximately holds for all the neural networks we study in this paper. Figure \ref{fig:weight-linearity} shows a test of the assumption on our compressed computation model, by checking whether inactive components in the APD decomposition can be ablated in random combinations without substantially affecting the end result.

Whether the assumption of parameter linearity is satisfied by all the non-toy neural networks that we might want to decompose is ultimately an empirical question. Current theoretical frameworks for computing arbitrary circuits in superposition \citep{Vaintrob_Mendel_Kaarel_2024, Bushnaq_Mendel_2024} do seem to satisfy this assumption. They linearly superpose mechanisms in parameter space to perform more computations than the model has neurons\footnote{In such a manner that equation \ref{eq:ablation_curves} should be satisfied up to terms scaling as ca. $\mathcal{O}(\epsilon)$, where $\epsilon$ is the noise level in the outputs of the target model due to superposition}. This provides a tentative theoretical basis to think that real models using computation in superposition do the same.

\tocless\subsection{Deriving the losses used in APD from the Minimum Description Length Principle}\label{app:MDL}

\tocless\subsubsection{Idealised loss: Minimum description length loss, $\mathcal{L}_{\text{MDL}}$}
We suspect that many models of interest only use a fraction of the mechanisms they have learned to process any particular input. Thus, we want a decomposition of our models into components such that the total complexity of the components used on any given forward pass (measured in bits) is minimized.

\paragraph{Motivating case: Parameter components that are rank 1 and localized in one layer. }
Suppose the elementary components in our model were all very simple, with each of them being implemented by a rank $1$ weight matrix $P_{c,l,i,j}= U_{c,l,i} V_{c,l,j}$ in some layer $l$ of the network.\footnote{We think that the total number of components $C$ here seems in theory capped to stay below the total number of network parameters $C=\mathcal{O}(N)$. See \cite{Bushnaq_Mendel_2024} for discussion.}
If we wanted to minimize the complexity used to describe the model's behavior on a given data point $x$, then we should minimize the number of components that have a non-zero causal influence on the output on that data point. In other words, we want to optimise the component attributions $A(x)$ to be sparse.

With a dense code, the attributions $A_c(x)$ on a given input $x$ would cost $\sum^C_{c=1} \alpha=C \alpha$ bits to specify, where $\alpha$ is the number of bits of precision we use for a single $A_c(x)$.

However, with a sparse code, we would instead need $\sum^C_{c=1} \vert\vert A_c(x)\vert\vert_0 \left(\alpha+\log_2(C)\right)$ bits, where $\vert\vert A_c(x)\vert\vert_0$ is the $L_0$ `norm' of $A_c(x)$. If the parameter component attributions $A_c(x)$ are sparse enough, this can be a lot lower than $C\alpha$. This leverages the fact that we can list only the indices and attributions of the subnets with non-zero $A_c(x)$. This requires $\log_2(C)$ bits for the index and $\alpha$ bits for the attribution.

\paragraph{General case: Parameter components that have arbitrary rank and may be distributed across layers.}
We do not expect all the parameter components of models to always be rank $1$ matrices; they may be arbitrary rank and span multiple layers. 
\footnote{Nevertheless, current hypotheses for how models might implement computations in superposition suggest that components would tend to be low-rank. \citep{Bushnaq_Mendel_2024}. Otherwise, there would just not be enough spare description length to fit all the (high rank) parameter components that are necessary to do the computation in superposition into the network.}

We can treat this similar to the motivating case above, but where a parameter component that consists of a rank $2$ matrix can be represented as two rank $1$ matrices that always co-activate. 
If two rank $1$ matrices almost always coactivate, then we can describe their attributions in two ways:
\begin{enumerate}
    \item If we consider them as \textbf{two separate components}, then we would need $2\log_2(C)+2\alpha$ bits to describe their attributions for each data point they activate on ($\log_2(C)$ for the index and $2\alpha$ for the two attributions). 
    \item However, if we consider them as \textbf{one separate component}, then we only need one index to identify both of them, and therefore only need $\log_2(C)+2\alpha$ bits
\end{enumerate}

This means that we may be able to achieve shorter description lengths using a mixed coding scheme that allows for both dense and sparse codes. Thus, if we use a mixed coding scheme that allows rank $1$ parameter components to be aggregated into higher dimensional components, it gives us a description length of
\begin{align}
\mathcal{L}_{\text{MDL}}(x)& =\sum^C_{c=1} \vert\vert A_c(x)\vert\vert_0 \left(\alpha \sum_l \text{rank}(P_{c,l})+\log_2(C)\right) \\
& = \log_2(C) \left(  \sum^C_{c=1} \vert\vert A_c(x)\vert\vert_0 \right) + 
\alpha \left(  \sum^C_{c=1} \vert\vert A_c(x)\vert\vert_0  \sum_l \text{rank}(P_{c,l}) \right) \\
& =: \log_2(C) \mathcal{L}_{\text{minimality}}^{\text{idealized}}(x) + \alpha \mathcal{L}_{\text{simplicity}}^{\text{idealized}}(x)
\end{align}
where $\sum_l\text{rank}(P_{c,l})$ is the total rank of component $P$ summed over the weight matrices in all the components of the network. 

Optimizing our components $P_c$ to minimize $\mathcal{L}_{\text{MDL}}(x)$ would then yield a decomposition of the network that uses only small values for the total number of active components
and the total rank of the active components on a particular forward pass.

The prefactor $\alpha$ in this equation then sets the point at which two lower-rank components coactivate frequently enough that merging them into a single higher-rank component lowers the overall loss. 
Thus, $\alpha$ is effectively a hyperparameter controlling the resolution of our decomposition.
As $\alpha$ increases, the threshold for merging components rises, with all components becoming rank $1$ in the limit $\alpha\rightarrow \infty$. 
If we set $\alpha=0$, all components would merge, so our decomposition would simply return the target network's parameter vector.

\paragraph{Full idealised MDL loss}

For our the loss term that we use to train our parameter components,  we want a decomposition that approximately sums to the target parameters and minimises description length. We can accomplish this by adding a faithfulness loss
\begin{equation}\label{eq:faithfulness_loss}
\begin{aligned}
\mathcal{L}_{\text{faithfulness}}&=\sum_{l,i,j}{\left( \theta^{*}_{l,i,j}- \sum^C_{c=1} P_{c,l,i,j}\right)}^2\,,\\
\end{aligned}
\end{equation}
to our minimum description length loss. Our full loss is then:\footnote{Here, we've absorbed $\log(C)$ from $\mathcal{L}_{\text{MDL}}$ in the previous section into $\beta$.}
\begin{equation}\label{eq:loss_ideal}
\begin{aligned}
\mathcal{L}_{\text{faithfulness}}+\mathcal{L}_{\text{MDL}}(x)&= \mathcal{L}_{\text{faithfulness}}+\beta \mathcal{L}_{\text{minimality}}^{\text{idealized}}(x)+\alpha \mathcal{L}_{\text{simplicity}}^{\text{idealized}}(x)\\
\end{aligned}
\end{equation}
However, this idealized loss would be difficult to optimize since the $L_0$ `norm' $\vert\vert A_c(x)\vert\vert_0$ and $\text{rank}(P_{c})$ are both non-differentiable. We therefore must optimize a differentiable proxy of this loss instead. 

We have devised two different proxy losses for this, leading to two different implementations of APD. The first uses a \textbf{top-$k$ formulation} (Section \ref{app:top_k}), whereas the second assigns an $L_p$ penalty to attributions \ref{app:l_p_loss}. We primarily use the top-$k$ formulation in our work. But we include the $L_p$ version for explanatory purposes.

\tocless\subsubsection{Practical loss: Top-$k$ formulation of APD}\label{app:top_k}

\paragraph{Approximating $\mathcal{L}_{\text{minimality}}^{\text{idealized}}$ in the top-$k$ formulation}
We can approximate optimizing for the loss in equation \ref{eq:loss_ideal} with a top-$k$ approach: We run the network once on data point $x$ and collect attributions $A_c(x)$ for each parameter component $P_c$. Then, we select the parameter components with the top-$k$ largest attributions and perform a forward pass using only those components. 
\begin{equation}
\begin{aligned}
s_c(x) &\in \{0,1\}\,\\
s_c(x)&=\text{top-k}(\{A_c(x)\})\\
\kappa(x)&:= \sum^C_{c=1} s_c(x) P_c\\
\end{aligned}
\end{equation}
This `sparse' forward pass should ideally only involve the structure in the network that is actually used on this specific input, so it should give the same result as a forward pass using all $P_c$. We can optimise for this using a loss 
\begin{equation}\label{eq:loss_topk}
\begin{aligned}
\mathcal{L}_{\text{minimality}}(P\vert \theta^*, X)&= D\left(f(x\vert \theta^*),f(x\vert \kappa(x))\right)\,.\\
\end{aligned}
\end{equation}
where $D$ is some distance measure between network outputs, e.g. MSE loss or KL-divergence. Minimising $\mathcal{L}_{\text{minimality}}$ for a small $k$ then approximately minimises
$\sum^C_{c=1} \vert\vert A_c(x)\vert\vert_0$ in the ideal loss. 
\paragraph{Reconstructing hidden activations} It is possible that reconstructing the network outputs on the sparse forward pass is not a strong enough condition to ensure that the components we find correspond to the mechanisms of the network, particularly since our attributions are imperfect. To alleviate this, we can additionally require some of the model's hidden activations on the sparse forward pass to reconstruct the target model's hidden activations. This can also aid training dynamics in deeper models, as APD can match the target model layer by layer instead of needing to re-learn everything from scratch. However, theories of computation in superposition predict that unused components still contribute noise to the model's hidden preactivations before non-linearities, which is then filtered out \citep{hänni2024mathematicalmodelscomputationsuperposition,Bushnaq_Mendel_2024}. So we do not necessarily want to match the hidden activations of the target model everywhere in the network. Finding a principled balance in this case is still an open problem. We use a hidden activation reconstruction loss for our single and multilayer models of compressed computation in Section \ref{sec:residmlp_1layer} and Section \ref{sec:residmlp_2layer}.

\paragraph{Approximating $\mathcal{L}_{\text{simplicity}}^{\text{idealized}}$ in the top-$k$ formulation} To approximate $\mathcal{L}_{\text{simplicity}}^{\text{idealized}}$, we need some tractable objective function that approximately  minimises $\text{rank}(P_{c})$.  We use the \textit{Schatten norm}: The rank of a matrix $M$ can be approximately minimised by minimising $\vert\vert M\vert\vert_p$ \citep{giampouras2020novelvariationalformschattenp} with $p\in(0,1)$:
\begin{equation}
\vert\vert M\vert\vert_p:=\left(\sum_m \vert\lambda_m \vert^p\right)^{\frac{1}{p}}
\end{equation}
Here, $\lambda_m$ are the singular values of $M$. So, we can approximate $\text{rank}(P_c)$ in the loss with 
\begin{equation}
\sum^C_{c=1}\vert\vert P_c\vert\vert^p_p =\sum^C_{c=1}\sum_{l,m}\vert\lambda_{c,l,m} \vert^p\,,
\end{equation}
where $\lambda_{c,l,m}$ is singular value $m$ of component $c$ in layer $l$.

Performing a singular value decomposition for every component at every layer every update step would be cumbersome. We can circumvent this by parametrizing our components in factorized form, as a sum of outer products of vectors $U,V$:
\begin{equation}\label{eq:factorized}
P_{c,l,i,j}:= \sum_k U_{c,l,m,i} V_{c,l,m,j}
\end{equation}
If we now replace $\lambda_{c,l,m}$ with 
\begin{equation}
\lambda_{c,l,m}\rightarrow \left(\sum_{i,j} U^2_{c,l,m,i} V^2_{c,l,m,j}\right)^{\frac{1}{2}}
\end{equation}
then $V_c$ and $U_c$ will be incentivised to effectively become proportional to the right and left singular vectors for subnet $P_c$. 

The Schatten norm of $P_c$ can then be written in factorized form as:
\begin{equation}\label{eq:simplicity_loss}
\begin{aligned}
\mathcal{L}_{\text{simplicity}}(x)&=\sum^C_{c=1} s_c(x)\sum_{l,m}\left(\sum_{i,j} U^2_{c,l,m,i} V^2_{c,l,m,j}\right)^{\frac{p}{2}}\,.
\end{aligned}
\end{equation}
\paragraph{Full set of loss functions in the top-$k$ formulation}
To summarise, our full loss function is
\begin{equation}
\begin{aligned}
\mathcal{L}(x)&= \mathcal{L}_{\text{faithfulness}}+\beta \mathcal{L}_{\text{minimality}}(x)+\alpha \mathcal{L}_{\text{simplicity}}(x)\\
\mathcal{L}_{\text{faithfulness}}&=\sum_{l,i,j}{\left( \theta^{*}_{l,i,j}- \sum^C_{c=1} P_{c,l,i,j}\right)}^2\\
\mathcal{L}_{\text{minimality}}(x)&=D\left(f(x\vert \theta^*),f(x\vert \sum^C_{c=1} s_c(x) P_{c})\right)\\
\mathcal{L}_{\text{simplicity}}(x)&=\sum^C_{c=1} s_c(x)\sum_{l}\vert\vert P_c\vert\vert^p_p\,.
\end{aligned}
\end{equation}
The components $P_c$ are parametrised as
\begin{equation}
P_{c,l,i,j}:= \sum_k U_{c,l,m,i} V_{c,l,m,j}\,.
\end{equation}
The top-$k$ coefficients $s_c(x)$ are chosen as
\begin{equation}
\begin{aligned}
s_c(x) &\in \{0,1\}\,\\
s_c(x)&=\text{top-k}(\{A_c(x)\})\\
\end{aligned}
\end{equation}
where $A_c(x)$ are attributions quantifying the effect of components $P_c$ on the network, computed with attribution patching as in equation \ref{eq:attribution}, or with some other attribution method. Finally, $\vert\vert P_c\vert\vert_p$ denotes the Schatten norm, and $p\leq 1.0$ is a hyperparameter. 

$\mathcal{L}_{\text{minimality}}(x)$ may include additional terms penalizing the distance $D$ between some of the hidden activations of the target model $\theta^*$, and the sparse forward pass using parameters $\sum^C_{c=1} s_c(x) P_{c}$.

We use batch top-$k$ instead of top-$k$ \citep{bussmann2024batchtopk}, picking the components with the largest attributions over a batch of datapoints instead of single inputs.

\tocless\subsubsection{Alternative practical loss: APD formulation that uses an $L_p$ penalty on attributions}\label{app:l_p_loss}

As an alternative to the top-$k$ loss, we can also approximately optimize for loss \ref{eq:loss_ideal} with an  $L_p$ approach. Optimizing the $L_p$ norm with $p\leq 1$ will tend to yield solutions with small $L_0$ `norm', while still being differentiable. 
So we can replace $\vert\vert A_c(x)\vert\vert_0$ in the loss with $\vert A_c(x)\vert^p$. Our losses would then be
\begin{equation}
\begin{aligned}
\mathcal{L}_{\text{minimality}}^{L_p}(x)&=\sum^C _{c=1} \vert A_c(x)\vert^{p_1} \\
\mathcal{L}_{\text{simplicity}}^{L_p}(x)&=\sum^C_{c=1}\sum_l\vert A_c(x)\vert^{p_1} \,\left(\sum_{i,j} U^2_{c,l,m,i} V^2_{c,l,m,j}\right)^{\frac{p_2}{2}}\,,
\end{aligned}
\end{equation}
where $p_1,p_2\leq 1.0$ are the $p$-norms of the attributions and the Schatten norm of the matrices respectively.

\begin{figure}[h!] 
    \centering
    \includegraphics[width=1.0\linewidth]{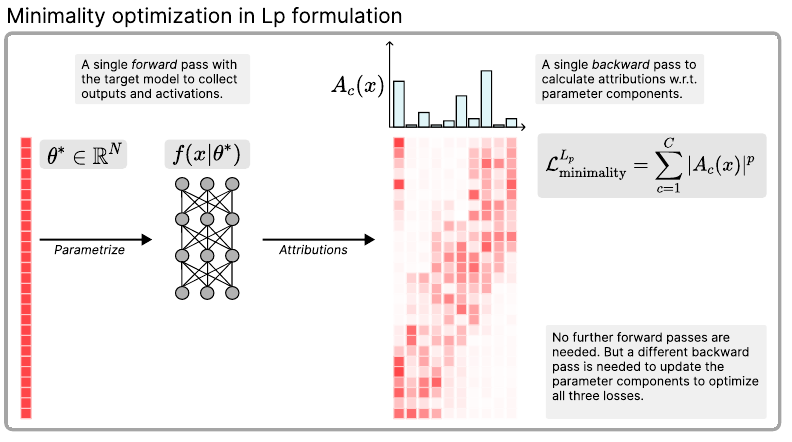}
    \caption{Optimizing $\mathcal{L}_{\text{minimality}}^{L_p}$}\label{fig:minimality_step_l_p}
\end{figure}

We did not thoroughly explore this implementation because our early explorations that used the $L_p$ approach did not work as well as our top-$k$ implementation for unknown reasons. We may revisit this approach in the future.

\section{Further experiments}

\tocless\subsection{Hand-coded gated function model: Another cross-layer distributed representation setting}\label{app:handcoded_gated_model} %

\begin{figure}
    \centering
    \includegraphics[width=1\linewidth]{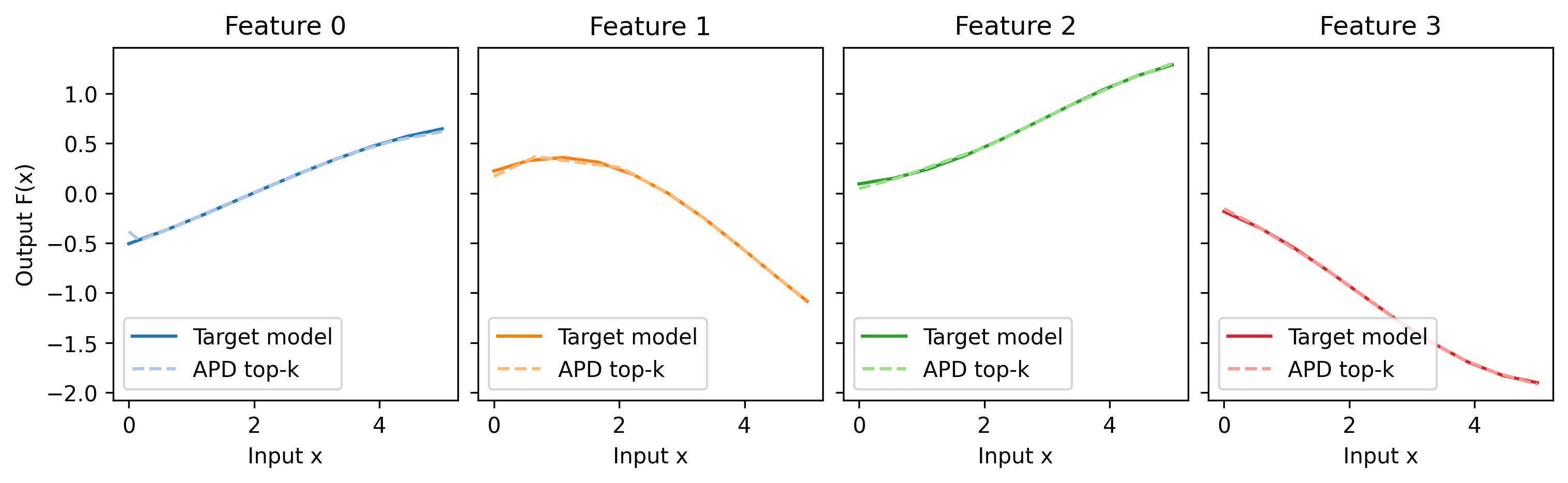}
    \caption{
        Hand-coded gated function model: The four functions $f_i(x)$ implemented by the
        hand-coded gated function model (solid lines), and the outputs
        of the top-$k$ forward pass of the APD-decomposed
        model (dashed lines). The APD model almost perfectly
        matches the hand-coded network.
    }
    \label{fig:piecewise_model_functions_paper}
\end{figure}

\tocless\subsubsection{Setup}
In this task, we hand-code a target network to give an approximation to the sum of a set of trigonometric functions, governed by a set of control bits. The functions being approximated are of the form $F_i(x)= a_i \cos(b_i x + c_i) + d_i \sin(e_i x + f_i)+h_i$ with randomly generated coefficients $\{a_i,b_i,c_i,d_i,e_i,f_i,g_i\}$ drawn from uniform distributions (Table \ref{tab:coeff_distributions}) for each unique function $i$.

The input to the network is a vector, $(x, \alpha_0, \cdots, \alpha_{n-1})$, whose entries are a scalar $x\in[0, 5]$ and a set of $n$ binary control bits
$\alpha_i\in\{0,1\}$. The control bits $\alpha_i$ are sparse, taking a value of $1$ with probability $p=0.05$ and $0$ otherwise. A function is only "active" (i.e. it should be summed in the output of the network) when its corresponding control bit is on.

Similar to our model of cross-layer distributed representations in Section \ref{sec:residmlp_2layer}, we use $2$-layer residual MLP network with ReLU activations. This model is hand-crafted to have $n$ clearly separable mechanisms that each approximate a unique trigonometric function. Notably, each function is computed by a unique set of neurons.

The output of the target model is a piecewise approximation of $y(x) = \sum_i \alpha_i F_i(x)$ with $n$ functions $y_i(x)$.

\begin{table}[h!]
\centering
\begin{tabular}{|c|c|}
\hline
\textbf{Coefficient} & \textbf{Range} \\ \hline
\(a\)                & \(\mathcal{U}(-1, 1)\) \\ \hline
\(b\)                & \(\mathcal{U}(0.1, 1)\) \\ \hline
\(c\)                & \(\mathcal{U}(-\pi, \pi)\) \\ \hline
\(d\)                & \(\mathcal{U}(-1, 1)\) \\ \hline
\(e\)                & \(\mathcal{U}(0.1, 1)\) \\ \hline
\(f\)                & \(\mathcal{U}(-\pi, \pi)\) \\ \hline
\(g\)                & \(\mathcal{U}(-1, 1)\) \\ \hline
\end{tabular}
\caption{Ranges of coefficients sampled from uniform distributions for the functions used in the hand-coded gated function model.}
\label{tab:coeff_distributions}
\end{table}

 In our experiments, we use a total of $n=4$ unique functions, with each function using $m=10$ neurons to piecewise-approximate the functions $F_i(x)$.
 We show these approximated
 functions in Figure \ref{fig:piecewise_model_functions_paper} (solid lines).
 The 5 inputs of our network ($x$, $\alpha_0$, $\alpha_1$, $\alpha_2$, $\alpha_3$) are stored in the first 5 dimensions of the residual stream, alongside a dimension that we read off as the output of the network ($\hat{y}(x)$).
 To hand-code the piecewise approximation of the individual functions $y_i$ we
 randomly select $m$ neurons from the MLPs, typically distributed across layers. This also means that the value of $\hat{y}_i$ is not represented in the intermediate layers, but only in the final layer.

 We show the weights of the hand-coded target network in the leftmost panel of Figure \ref{fig:piecewise_subnetworks_graph_plots}.
 The graph shows the residual MLP network, with weights shown as lines. Each neuron is monosemantic, that is, it is used to approximate
 one of the $F_i(x)$ functions. Each neuron connects to the respective
 control bit $\alpha_i$ as well as the $x$ input. All neurons write
 to the output activation, which is the last dimension in the residual stream. The line color in Figure \ref{fig:piecewise_subnetworks_graph_plots} indicates which task
 (i.e. which function $F_i$) the weight implements; the line width
 indicates the magnitude of the weight.

When applied to this network, APD should partition the network weights $\theta^*$ into $C=n$ parameter components $P_c$, each corresponding to the weights for one approximated function $F_i(x)$ (i.e. of one colour).

\tocless\subsubsection{Results}
\begin{figure}
    \centering
    \includegraphics[width=1\linewidth]{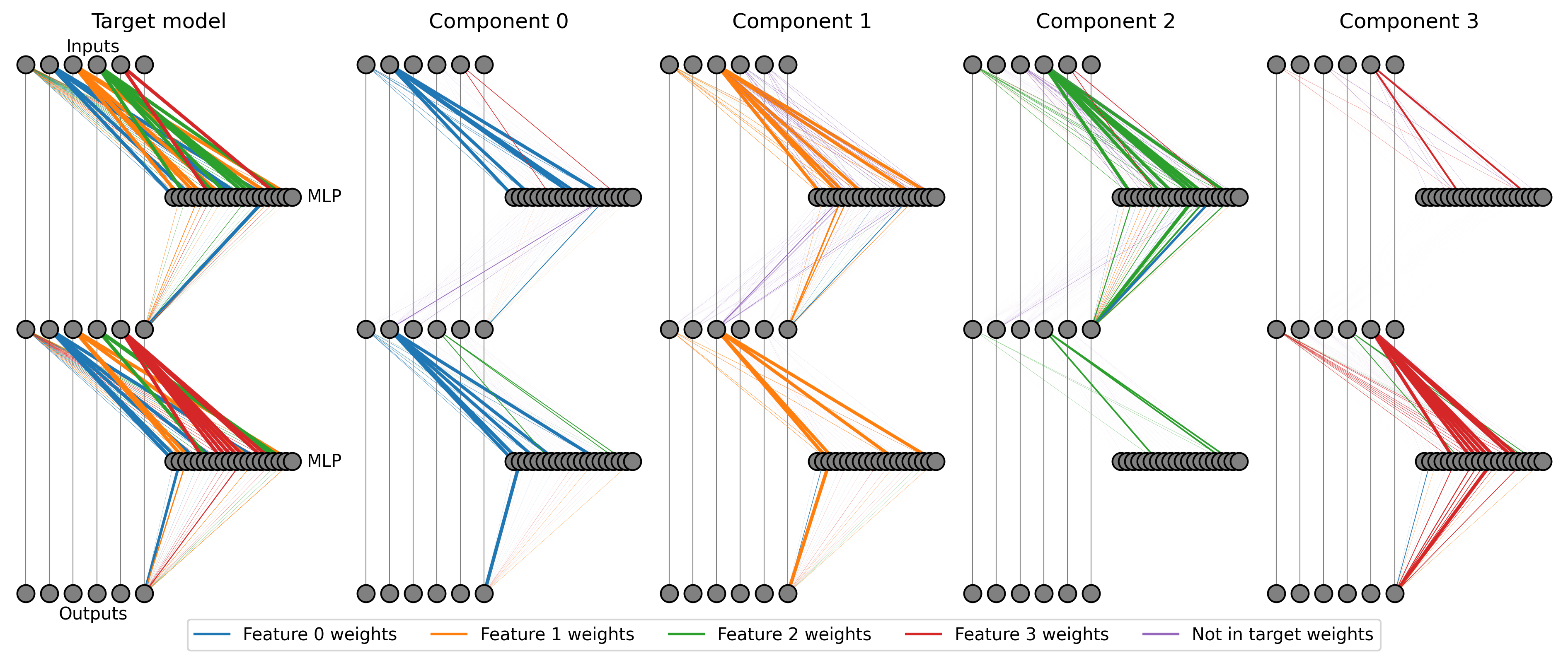}
    \caption{The parameters of the hand-coded gated function model decomposed into parameter components. \textbf{Leftmost panel:} The hand-coded network parameters, colored by the unique functions $F_i(x)$. \textbf{Other panels:} The parameter components identified by APD, coloured by the function they correspond to in the target model, or purple if the weight is zero in the target model.}
    \label{fig:piecewise_subnetworks_graph_plots}
\end{figure}

We find that APD can decompose this network into approximately correct parameter components. However, APD is particularly difficult to train in this setting, with only minor changes in hyperparameters causing large divergences. We hypothesize that this may be due to the fact that the ground truth network is itself hand-coded, not trained. We show a cherry-picked example (out of many runs that vary the number of MLP layers and number of functions) in Figure \ref{fig:piecewise_subnetworks_graph_plots}.

Figure \ref{fig:piecewise_subnetworks_graph_plots} shows the target network weights (leftmost column), and their decomposition into the four APD-generated components (remaining columns). We color the weights by which feature they correspond to in the target
model, or purple if the weight is not present in the target model.
We observe that the components mostly capture one function each
(most weights within a parameter component are the same color).

However, the solution is not perfect. Some weights that are not present in the target network are nevertheless nonzero in some of the parameter components. %
Additionally, the $W_{\rm out}^1$ weights of parameter component $2$ and $W_{\rm out}^0$ weights of parameter component $3$ seem to be absorbed into other parameter components. This may be due to the difficulty in training APD on a handcoded model as mentioned earlier, or may be a symptom of the simplicity loss $\mathcal{L}_{\text{simplicity}}$ not being fully layer-independent, causing an over-penalization of weights being in a layer on their own (see Appendix \ref{app:non-privileging} for a discussion on layer-privileging).

\section{Further analyses}\label{app:further_analyses}

\tocless\subsection{Analysis of the compressed computation target model}\label{app:resid_mlp_target}
In this section we provide more details about the performance
of the target residual MLP model that is used to train APD, as discussed in Section \ref{sec:residmlp_1layer}.

Recall that we train the target network to approximate $y_i = x_i + \text{ReLU}(x_i)$. Note that the model output can be written as
\begin{gather*}
    \mathbf{y} = W_U W_E \mathbf{x} + W_U W_{\rm out} \text{ReLU}(W_{\rm in} W_E \mathbf{x}).
\end{gather*}
Since $W_E$ consists of random unit vectors and is not trained. Also, $W_U = W_E^T$. As a result, the first summand already approximates a noisy identity and the second summand mostly approximates the ReLU function.

\begin{figure}
    \centering
    \includegraphics[width=1\linewidth]{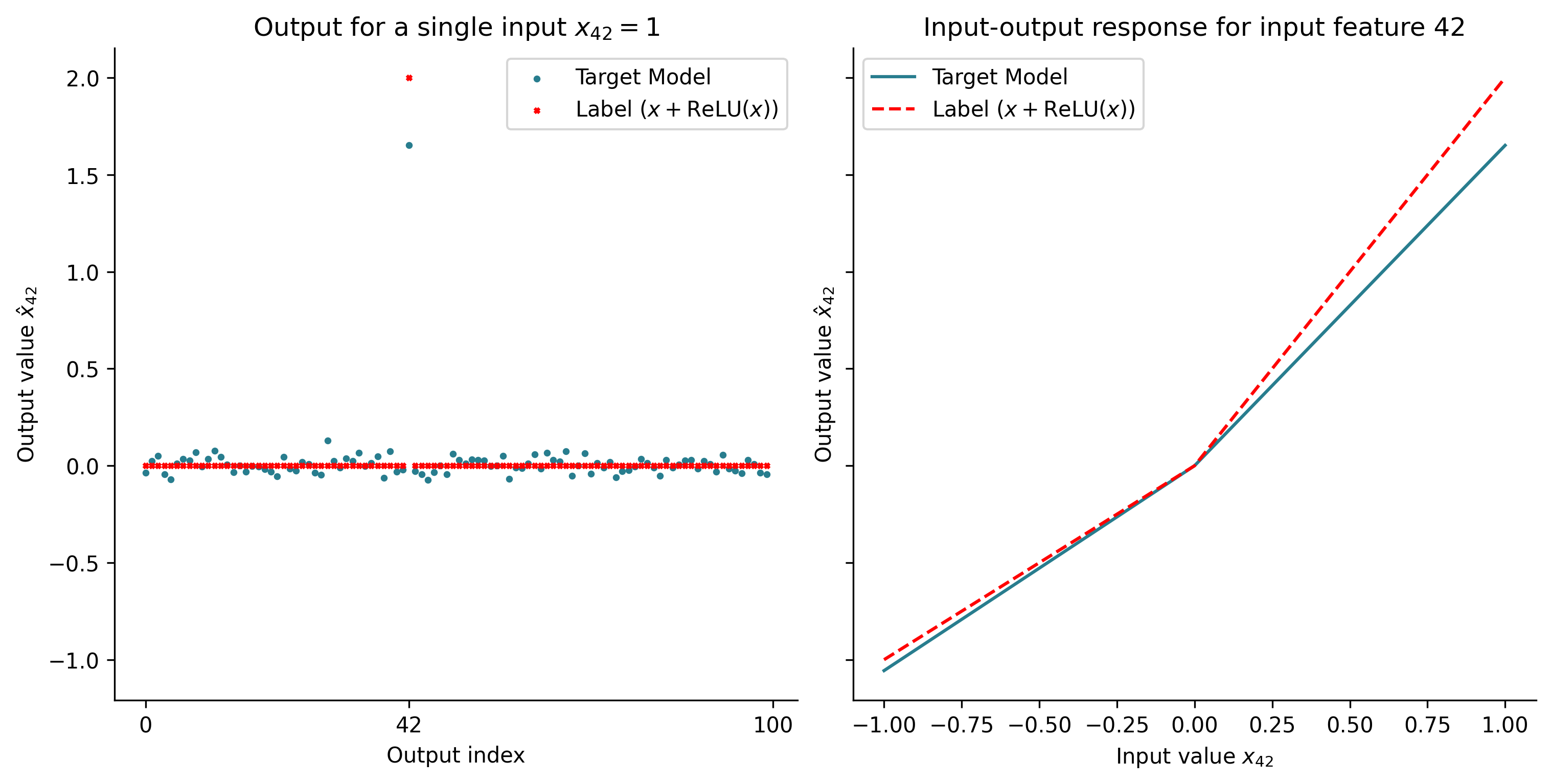}
    \caption{Output of the $1$-layer residual MLP target model compared to true labels for a single active input. \textbf{Left}: Output at all output indices for single one-hot input $x_{42}=1$. \textbf{Right}: Output at index $j=42$ for inputs with $x_{42}\in[0,1]$ and $x_j=0$ for $j\neq 42$.}
    \label{fig:app:residmlp_response_single}
\end{figure}

Figure \ref{fig:app:residmlp_response_single} (left)
shows the output of the model for an arbitrary one-hot
input ($x_{42}=1$). We see that the output
$\hat{x}_{42}\approx 1.6$ is close to the target
value of $2.0$, and the remaining outputs $\hat{x}_{i\neq 42}$
are close to $1.0$. We checked whether the noise in
the $\hat{x}_{i\neq 42}$ outputs comes from the $W_U W_E$
or MLP term, and found that it is dominated by the MLP
term.\footnote{This is not the case for small embedding
sizes, such as $d_{\rm resid}=100$. This is why we
chose a large embedding size to focus on the
MLP noise.}
We confirm that $\hat{x}_{42}$ indeed approximates a ReLU function for $\hat{x}\in[-1,1]$ in Figure
\ref{fig:app:residmlp_response_single} (right panel), though not perfectly. It appears to systematically undershoot the labels. We expect that this is due to the MSE loss: Although the model could scale the outputs (by scaling e.g. $W_{\rm out}$) to match $y_{42} = 2.0$, it would also increase the loss overall.

\begin{figure}
    \centering
    \includegraphics[width=1\linewidth]{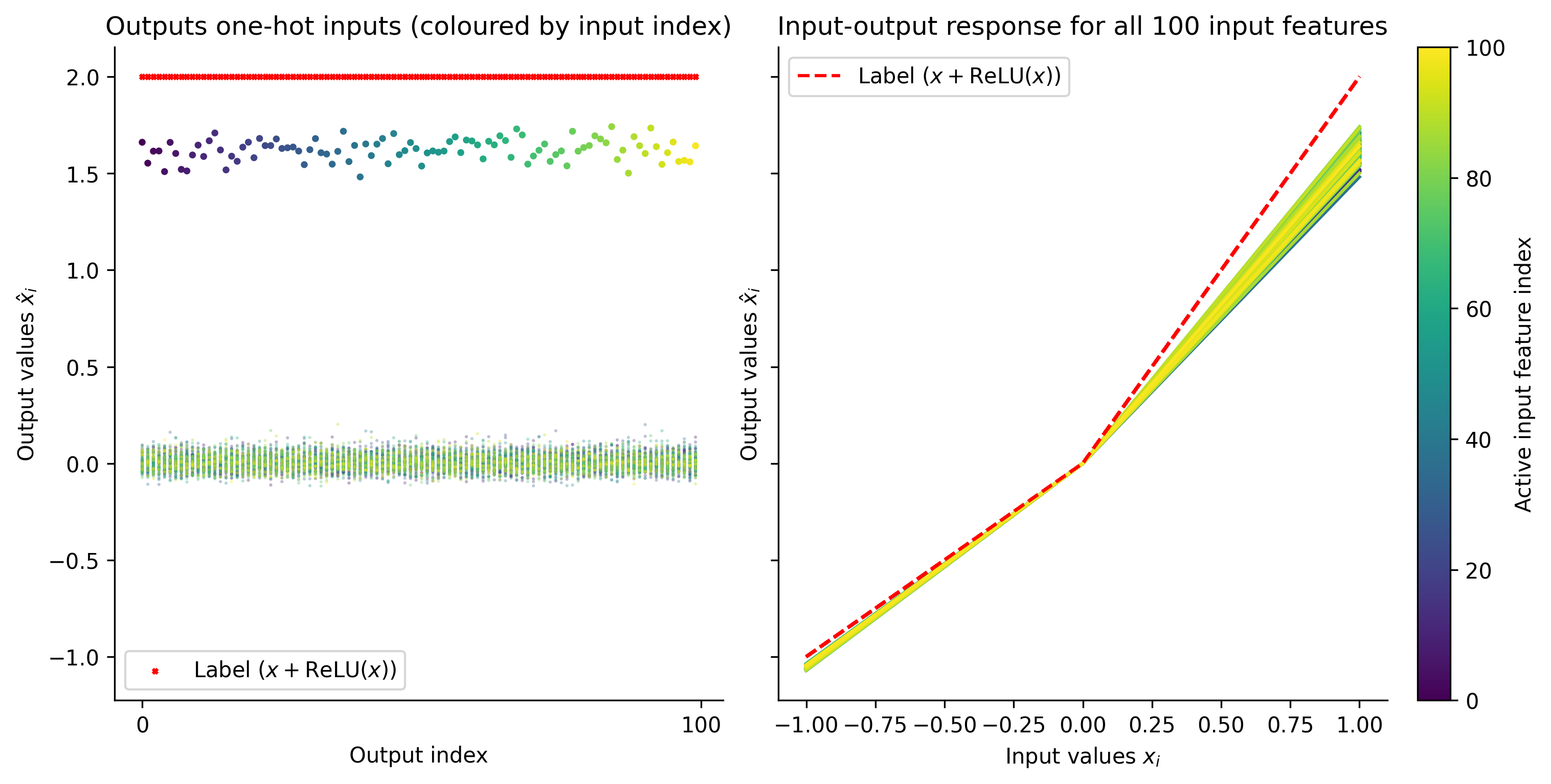}
    \caption{Output of the $1$-layer residual MLP target model compared to true labels for the full set of $100$ one-hot inputs. \textbf{Left}: Output at all output indices over the set of inputs. The point color indicates the active input feature, and label values are in red. \textbf{Right}: Output at index $i$ for inputs with $x_{i}\in[0,1]$ and $x_j=0$ for $j\neq i$. Line colors indicate the input feature index.}
    \label{fig:app:residmlp_response_multi}
\end{figure}

So far we have focused on the arbitrary input index $42$.
Figure
\ref{fig:app:residmlp_response_multi} repeats the same experiment but overlaying the results of all $100$ input features
(lines color indicating the input feature index). We can see that the
model treats all input features approximately the same.

\tocless\subsection{Analysis of the compressed computation APD model}\label{app:resid_mlp_1layer_apd}

For a setting like the compressed computation task, where the dataset consists of input features activating independently with probability $p$, a natural choice for the batch top-$k$ hyperparameter is a value close to $p$ multiplied by the number of input features. In our experiments, this would be $0.01\times100=1$. For this value of batch top-$k$ (and similar), there are batches in which APD must activate more parameter components than there are active features, and likewise, batches in which APD must activate fewer parameter components than there are active features. In our $1$-layer and $2$-layer residual MLP experiments in Section \ref{sec:residmlp_1layer} and Section \ref{sec:residmlp_2layer}, respectively, we chose the value of batch top-$k=1.28$ to be such that in almost no batches would there be more active input features than active components (we use a batch size of $256$).
The benefits of choosing this large batch top-$k$ value are:
\begin{enumerate}
    \item APD can learn to handle rarer samples with many active input features.
    \item Since there are very rarely more active input features than active components, the components are not encouraged to represent the computations of multiple input features.
\end{enumerate}

\begin{figure}
    \centering
    \includegraphics[width=1\linewidth]{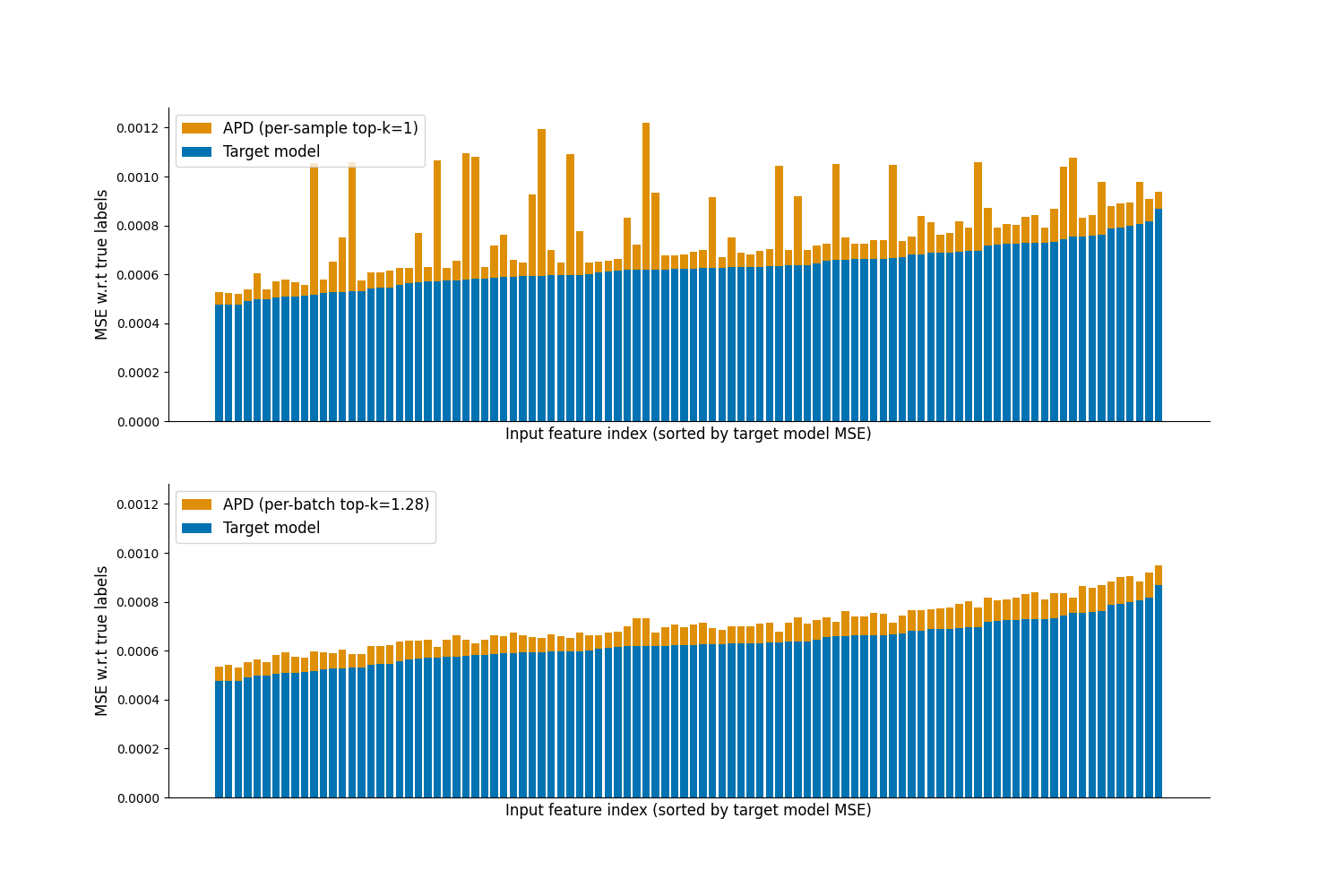}
    \caption{MSE for APD trained with batch top-$k=1.28$ in the $1$-layer residual MLP setting for samples with a single active input feature (i.e. one-hot), averaged over $100$k samples. \textbf{Top:} Comparison of the target model with the APD model when activating exactly one parameter component in each sample (i.e. top-$k=1$). \textbf{Bottom:} Comparison of the target model with the APD model using batch top-$k=1.28$. The batch top-$k$ mask is applied to the original training distribution and then samples without exactly one active input feature are filtered out.}
    \label{fig:app:resid_mlp_per_feature_performance_1layers}
\end{figure}

\begin{figure}
    \centering
    \includegraphics[width=1\linewidth]{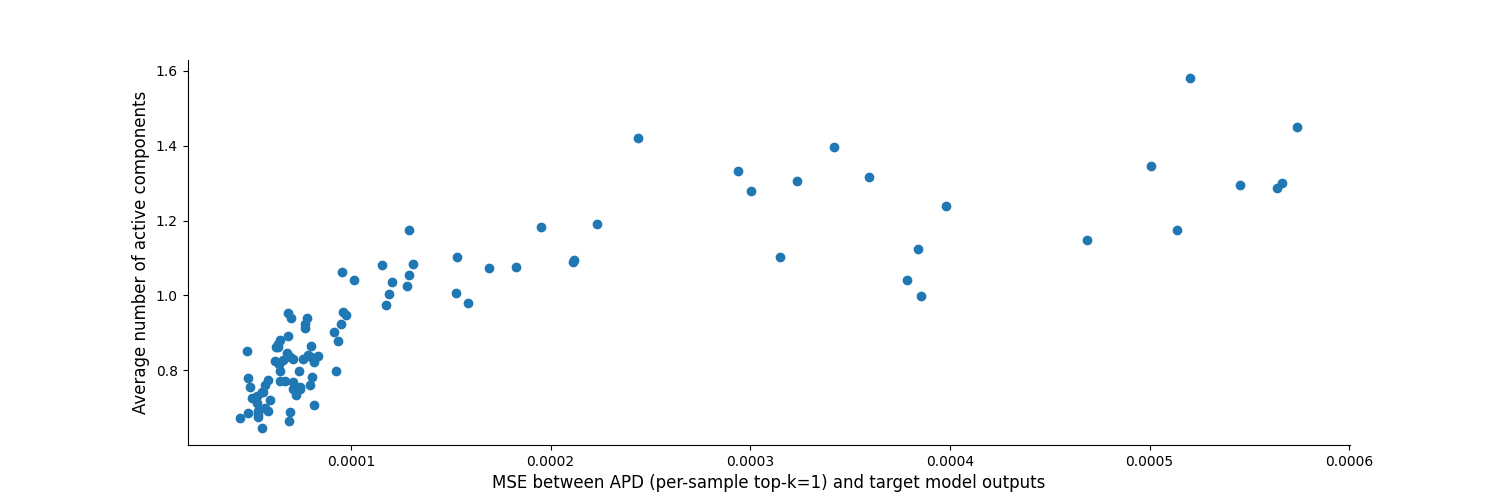}
    \caption{Relationship in the $1$-layer residual MLP setting between: (y-axis) the average number of active APD parameter components when using batch top-$k=1.28$, and (x-axis) the MSE between the target model outputs and the APD model when activating exactly one parameter component in each sample (i.e. top-$k=1$). MSE is measured only on samples with a single active input feature.}
    \label{fig:app:resid_mlp_avg_components_scatter_1layers}
\end{figure}

However, since there are extra active parameter components in most batches, APD exhibits a behavior where, for a subset of input features, it represents part of its computation in multiple parameter components. This phenomenon is illustrated in Figure \ref{fig:app:resid_mlp_per_feature_performance_1layers}, where the APD model achieves a low loss across all input features when using its trained batch top-$k=1.28$ setting (bottom). However, when constrained to activate only a single parameter component per sample, the model exhibits large losses for a non-negligible subset of the input features (top). These results are based on samples where only one input feature is active. This behavior is further characterized in Figure \ref{fig:app:resid_mlp_avg_components_scatter_1layers}. Samples with higher MSE loss under single-component activation tend to require more parameter components on the training distribution with batch top-$k=1.28$. 

\begin{figure}
    \centering
    \includegraphics[width=1\linewidth]{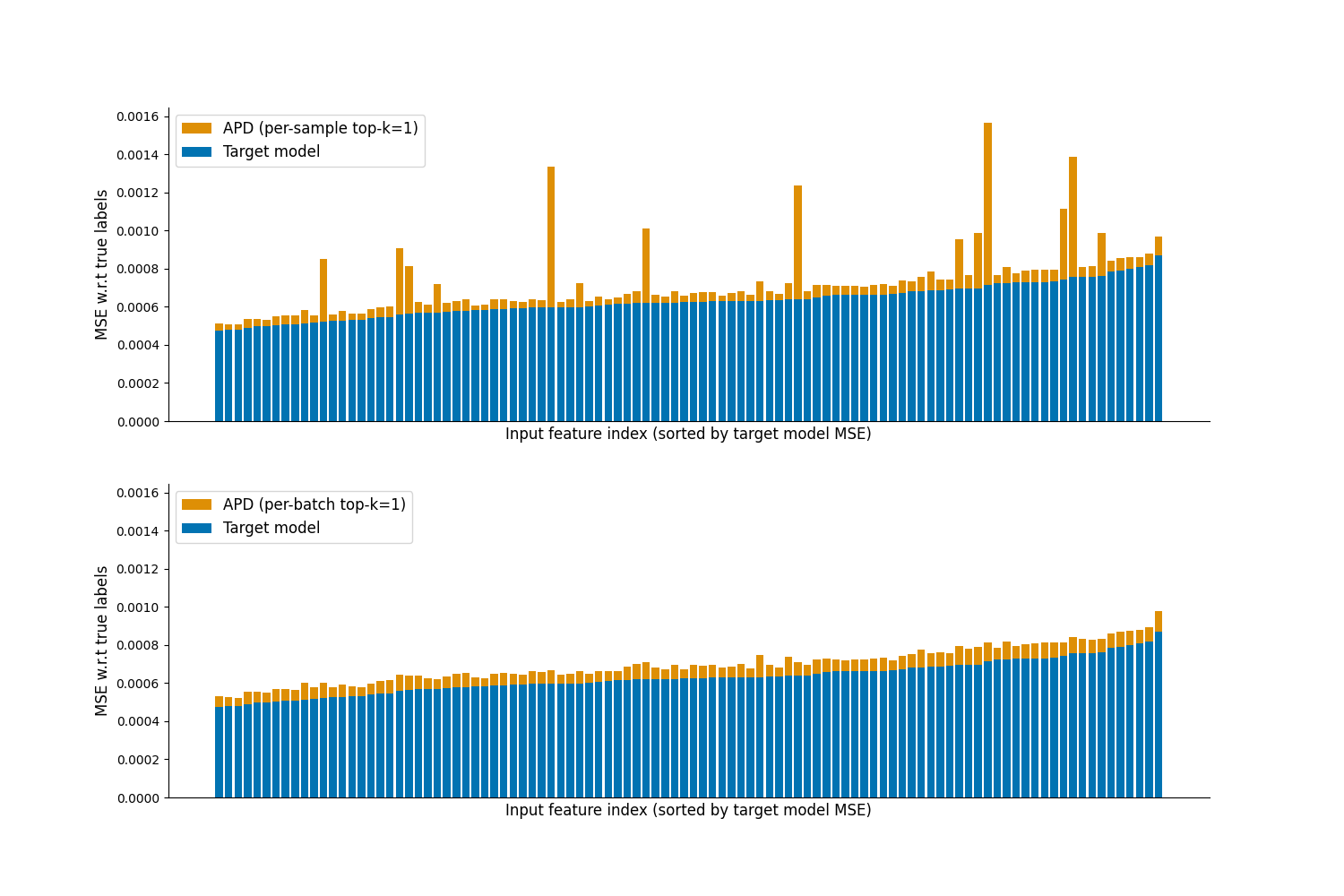}
    \caption{MSE for APD trained with batch top-$k=1$ in the $1$-layer residual MLP setting for samples with a single active input feature (i.e. one-hot), averaged over $100$k samples. \textbf{Top:} Comparison of the target model with the APD model when activating exactly one parameter component in each sample (i.e. top-$k=1$). \textbf{Bottom:} Comparison of the target model with the APD model using batch top-$k=1$. The batch top-$k$ mask is applied to the original training distribution and then samples without exactly one active input feature are filtered out.}
    \label{fig:app:resid_mlp_per_feature_performance_1layers_topk1}
\end{figure}

As shown in Figure \ref{fig:app:resid_mlp_per_feature_performance_1layers_topk1}, we see that training with a reduced batch top-$k$ value of $1$ (rather than $1.28$) reduces the number of input features that have a large MSE loss when only activating a single parameter component. However, the downside of using a smaller top-$k$ value is that we end up with more components that fully represent two different input feature computations, rather than one. See figures in the ``Toy Model of Compressed Computation (1 layer) with batch top-k$=1$'' section \href{https://api.wandb.ai/links/apollo-interp/h5ekyxm7}{here} for details. This should not be surprising; when top-$k$ is smaller, there are more batches in which the number of active input features is larger than the number of active components. APD is then incentivized to represent multiple input feature computations in a single parameter component to achieve a smaller $\mathcal{L}_{\text{minimality}}$ (though, at the cost of a larger $\mathcal{L}_{\text{simplicity}}$). 

\tocless\subsection{Analysis of the cross-layer distributed representations target model}\label{app:resid_mlp_2layer_target}

\begin{figure}
    \centering
    \includegraphics[width=1\linewidth]{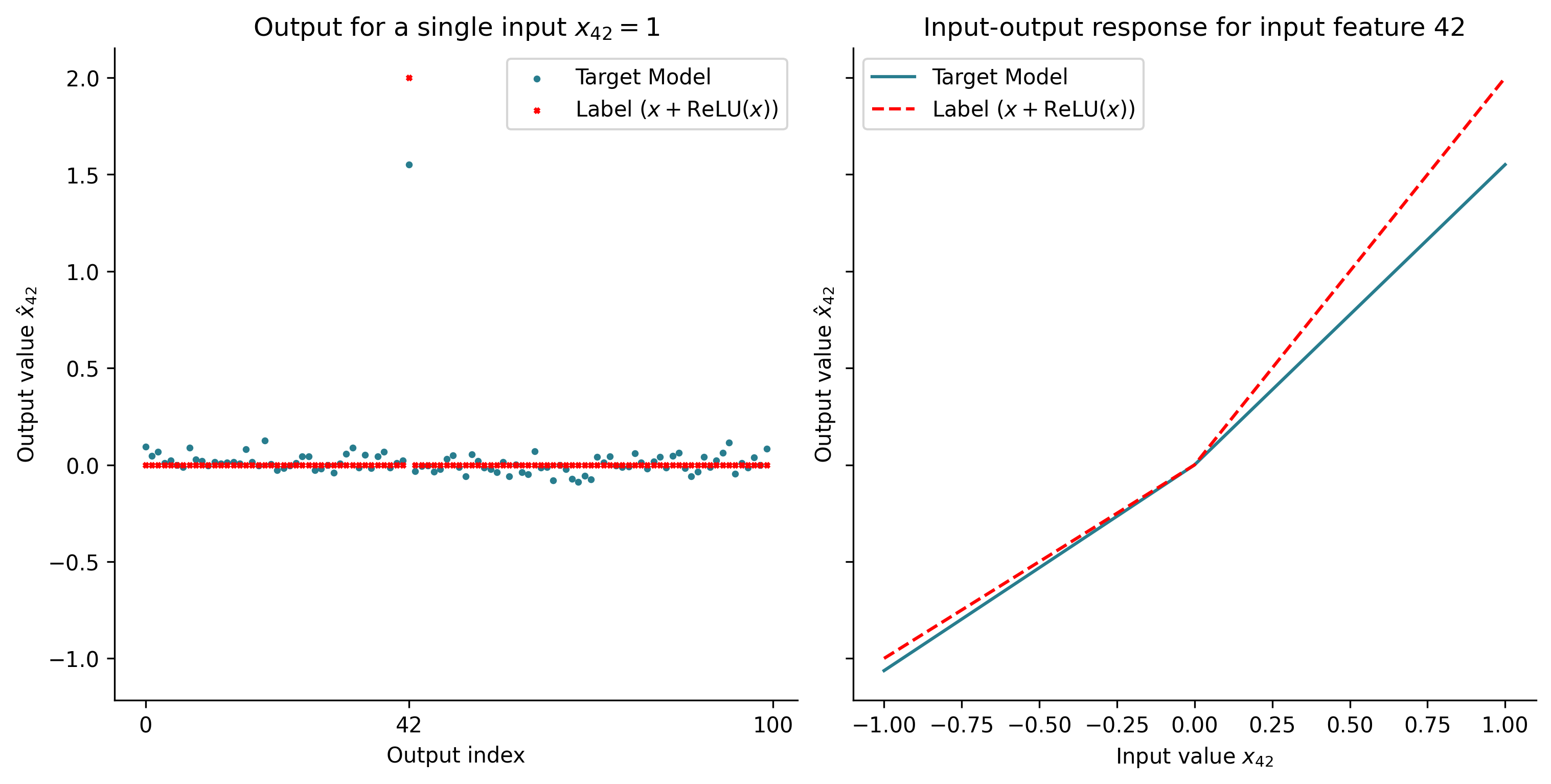}
    \caption{Output of the $2$-layer residual MLP target model compared to true labels for a single active input. \textbf{Left}: Output at all output indices for single one-hot input $x_{42}=1$. \textbf{Right}: Output at index $j=42$ for inputs with $x_{42}\in[0,1]$ and $x_j=0$ for $j\neq 42$.}
    \label{fig:app:residmlp_response_single_2layers}
\end{figure}

\begin{figure}
    \centering
    \includegraphics[width=1\linewidth]{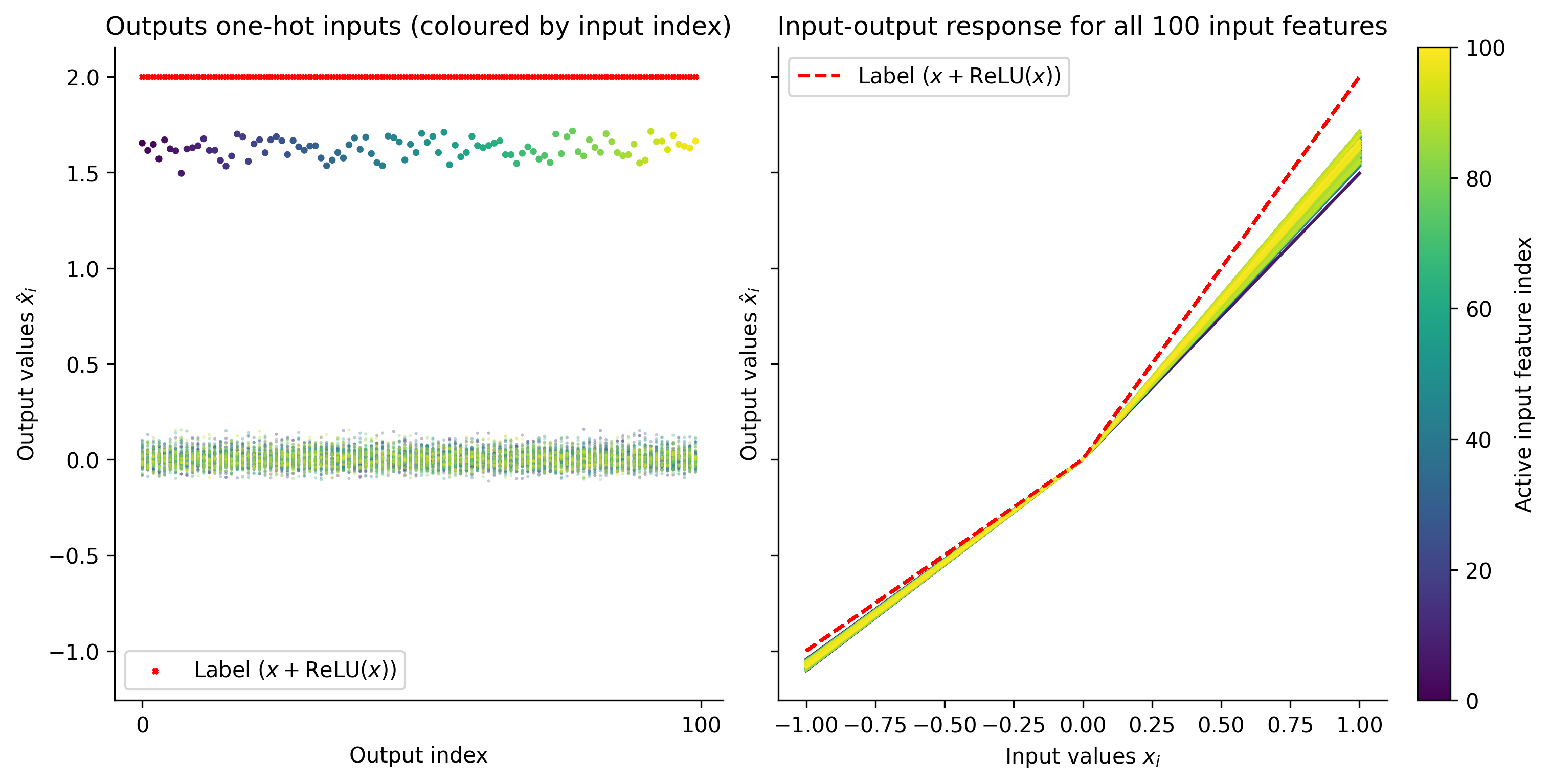}
    \caption{Output of the $2$-layer residual MLP target model compared to true labels for the full set of $100$ one-hot inputs. \textbf{Left}: Output at all output indices over the set of inputs. The point color indicates the active input feature, and label values are in red. \textbf{Right}: Output at index $i$ for inputs with $x_{i}\in[0,1]$ and $x_j=0$ for $j\neq i$. Line colors indicate the input feature index.}
    \label{fig:app:residmlp_response_multi_2layers}
\end{figure}

In Figures \ref{fig:app:residmlp_response_single_2layers} and \ref{fig:app:residmlp_response_multi_2layers}, we show that the trained target model for the cross-layer distributed representations setting in Section \ref{sec:residmlp_2layer} (i.e. $2$-layer residual MLP) is qualitatively similar to the target model in the compressed computation setting (i.e. $1$-layer residual MLP) we analyzed in Appendix \ref{app:resid_mlp_target}.

\tocless\subsection{Analysis of the cross-layer distributed representations APD model}\label{app:resid_mlp_2layer_apd}

Here, we show that the APD model for the cross-layer distributed representations setting in Section \ref{sec:residmlp_2layer} (i.e. $2$-layer residual MLP) is qualitatively similar to the APD model in the compressed computation setting (i.e. $2$-layer residual MLP) we analyzed in Section \ref{sec:residmlp_1layer}.

\begin{figure}
    \centering
    \includegraphics[width=1\linewidth]{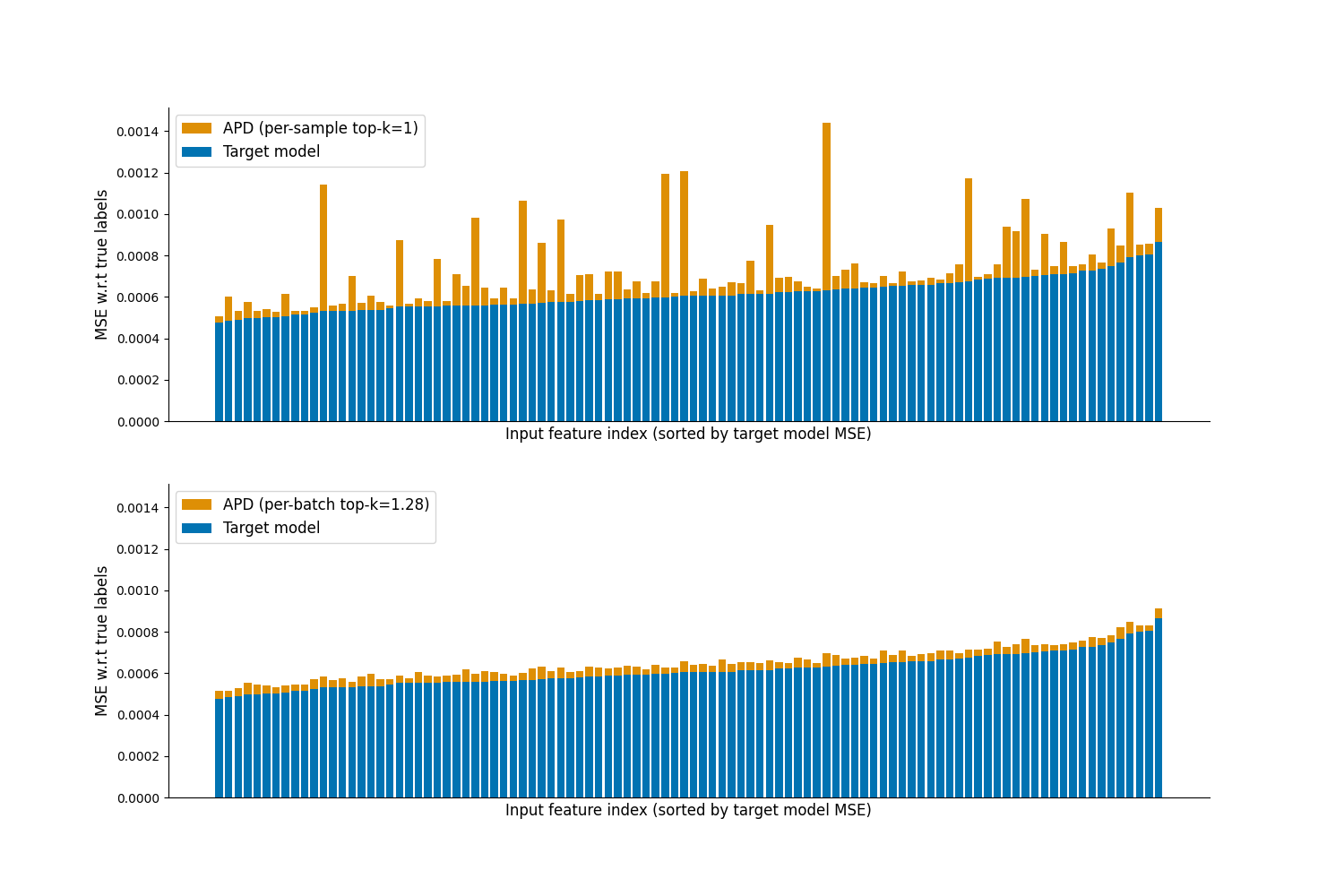}
    \caption{MSE for APD trained with batch top-$k=1.28$ in the $2$-layer residual MLP setting for samples with a single active input feature (i.e. one-hot), averaged over $100$k samples. \textbf{Top:} Comparison of the target model with the APD model when activating exactly one parameter component in each sample (i.e. top-$k=1$). \textbf{Bottom:} Comparison of the target model with the APD model using batch top-$k=1.28$. The batch top-$k$ mask is applied to the original training distribution and then samples without exactly one active input feature are filtered out.}
    \label{fig:app:resid_mlp_per_feature_performance_2layers}
\end{figure}

\begin{figure}
    \centering
    \includegraphics[width=1\linewidth]{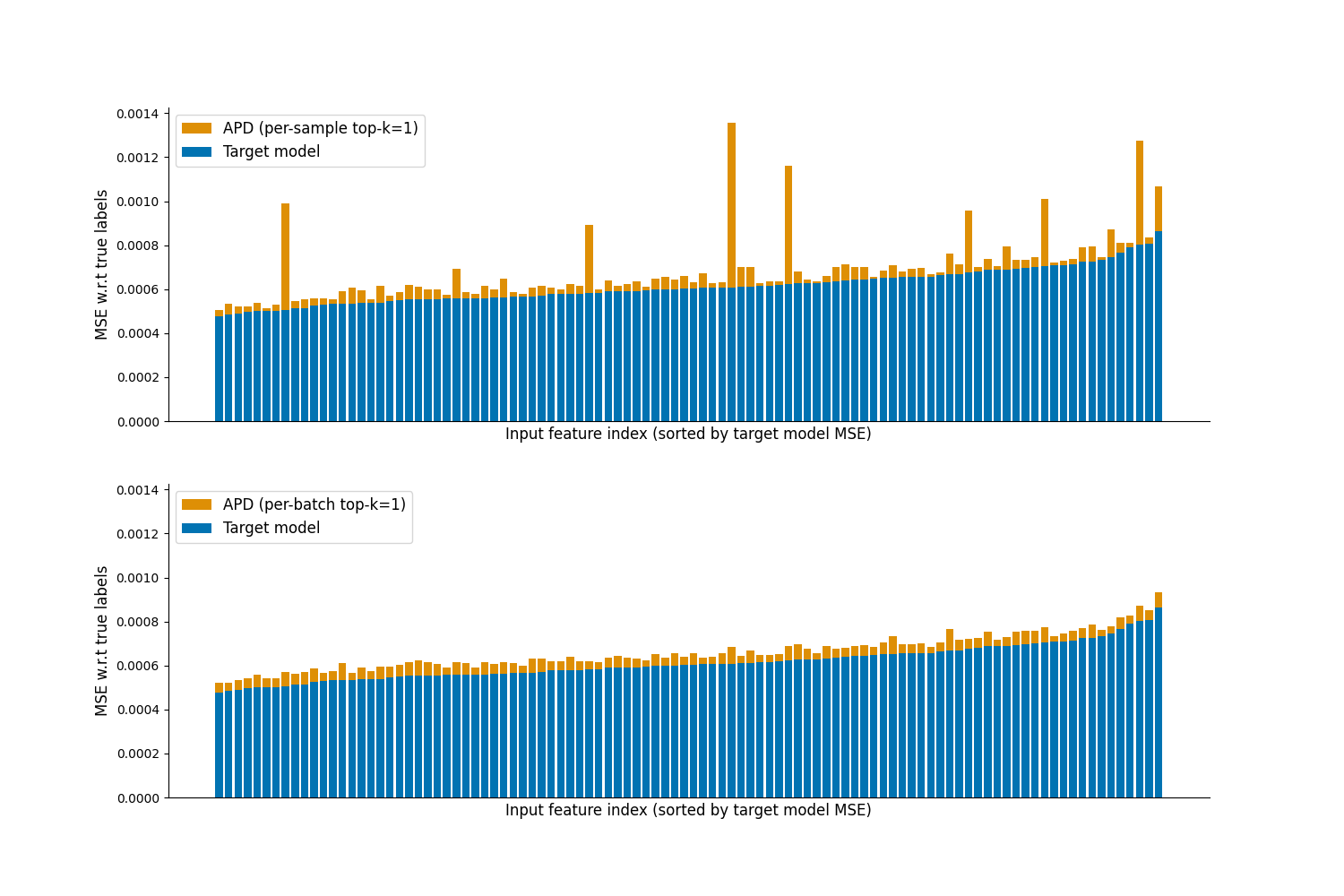}
    \caption{MSE for APD trained with batch top-$k=1$ in the $2$-layer residual MLP setting for samples with a single active input feature (i.e. one-hot), averaged over $100$k samples. \textbf{Top:} Comparison of the target model with the APD model when activating exactly one parameter component in each sample (i.e. top-$k=1$). \textbf{Bottom:} Comparison of the target model with the APD model using batch top-$k=1$. The batch top-$k$ mask is applied to the original training distribution and then samples without exactly one active input feature are filtered out.}
    \label{fig:app:resid_mlp_per_feature_performance_2layers_topk1}
\end{figure}

\begin{figure}
    \centering
    \includegraphics[width=1\linewidth]{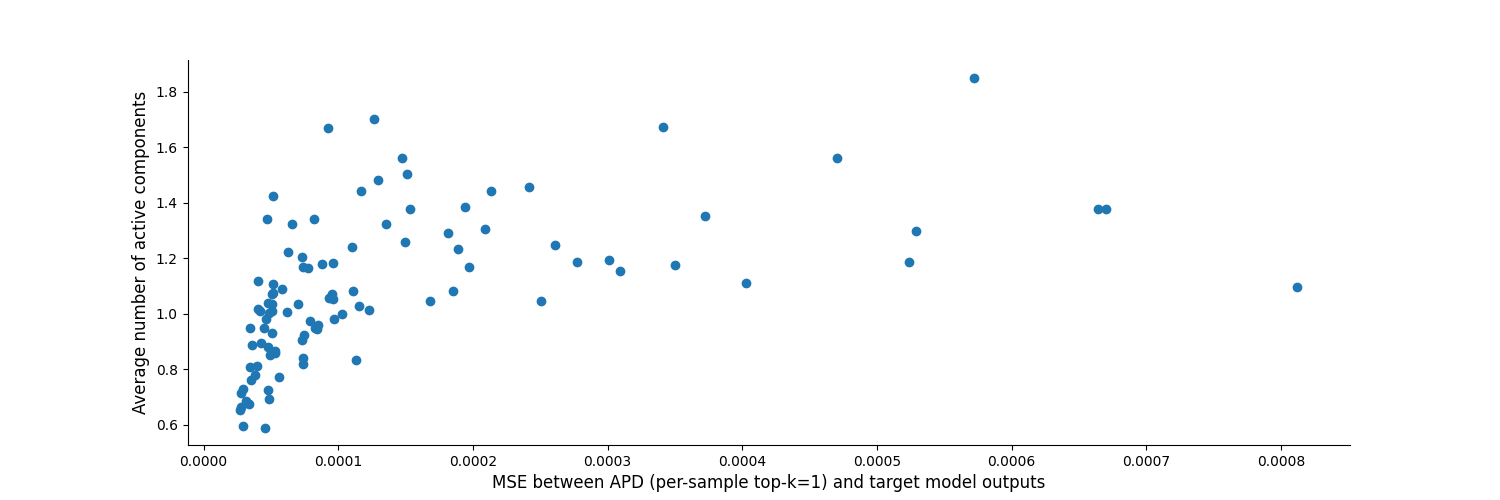}
    \caption{Relationship in the $2$-layer residual MLP setting between: (y-axis) the average number of active APD parameter components when using batch top-$k=1.28$, and (x-axis) the MSE between the target model outputs and the APD model when activating exactly one parameter component in each sample (i.e. top-$k=1$). MSE is measured only on samples with a single active input feature.}
    \label{fig:app:resid_mlp_avg_components_scatter_2layers}
\end{figure}

When running APD with batch top-$k=1.28$ in the $2$-layer residual MLP setting, we observe the same phenomenon previously described in Appendix \ref{app:resid_mlp_1layer_apd} for the $1$-layer case: certain input feature computations are not fully captured by individual parameter components (Figures \ref{fig:app:resid_mlp_per_feature_performance_2layers} and \ref{fig:app:resid_mlp_avg_components_scatter_2layers}). As in the $1$-layer setting, training with a reduced batch top-$k$ value of $1.28$ helps address this issue (Figure \ref{fig:app:resid_mlp_per_feature_performance_2layers_topk1}), though we again end up with more components that fully represent multiple input feature computations (see figures in the ``Toy Model of Compressed Computation ($2$ layer) with batch top-k$=1$'' section \href{https://api.wandb.ai/links/apollo-interp/h5ekyxm7}{here} for details).

It is worth noting that, if we instead enforce a rank-$1$ constraint on the parameter components in each network layer, we are able to get the best of both worlds. That is, APD does not learn parameter components that fully represent multiple input feature computations (it is unable to do this since this would require matrices with rank$>1$), and one is able to reduce the batch top-$k$ value to avoid having partial representations of an input feature computation across multiple components (in fact, leaving batch top-$k=1.28$ almost completely rectifies this issue in the rank-$1$). See the ``Rank-1 Toy Model of Cross-layer Distributed Representations ($2$ layers)'' section \href{https://api.wandb.ai/links/apollo-interp/h5ekyxm7}{here} for details.

\begin{figure}
    \centering
    \includegraphics[width=1\linewidth]{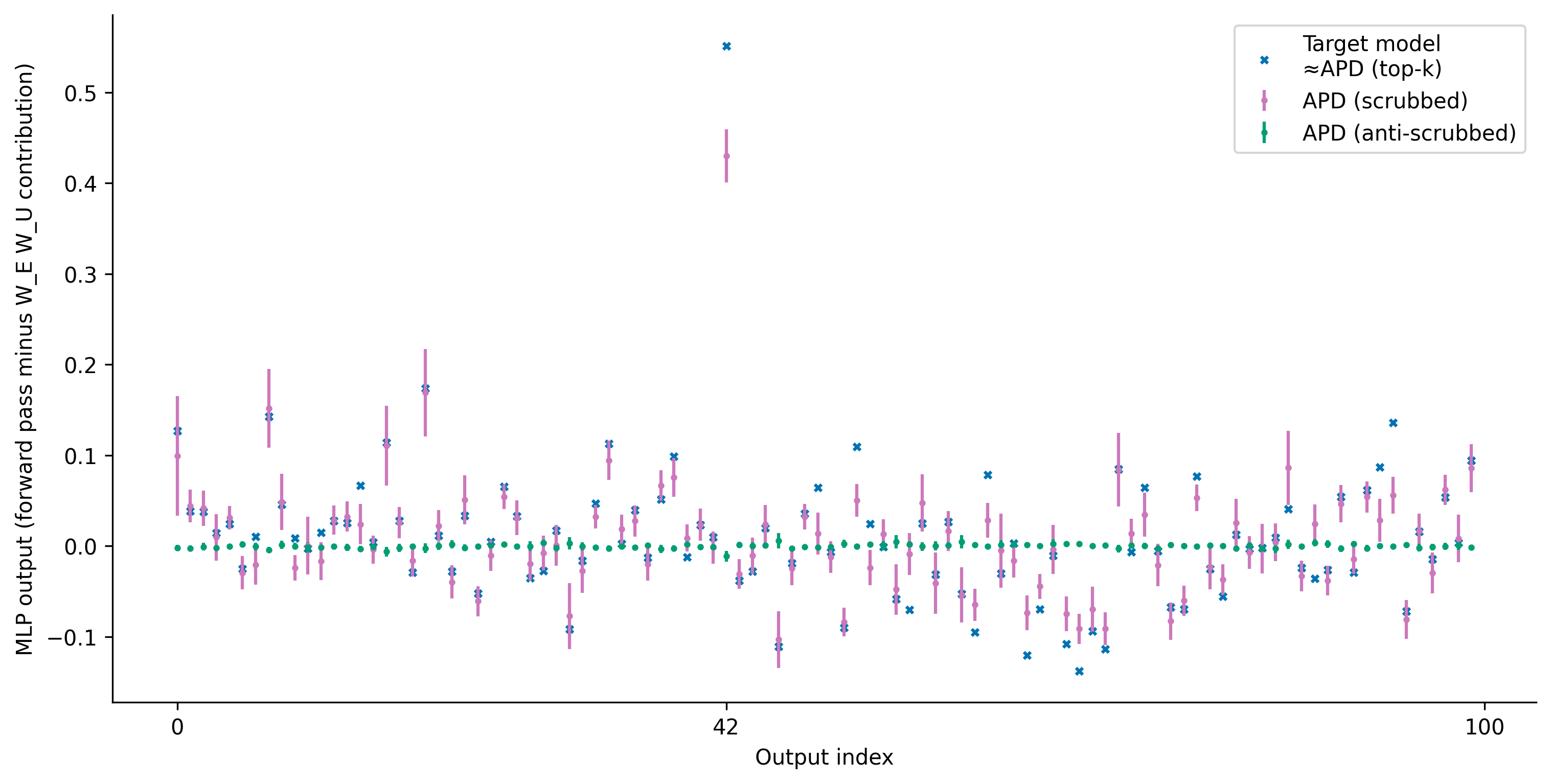}
    \caption{Output of multiple $2$-layer residual MLP APD forward passes with one-hot input $x_{42} = 1$ over $10$k samples, where half of the parameter components are ablated in each run. Purple lines show "scrubbed" runs (parameter component corresponding to input index $42$ is preserved), while green lines show "anti-scrubbed" runs (component $42$ is among those ablated). The target model output is shown in blue, which is almost identical to the output on the APD sparse forward pass (i.e. APD (top-$k$)).}
    \label{fig:weight-linearity-2layers}
\end{figure}

\begin{figure}
    \centering
    \includegraphics[width=1\linewidth]{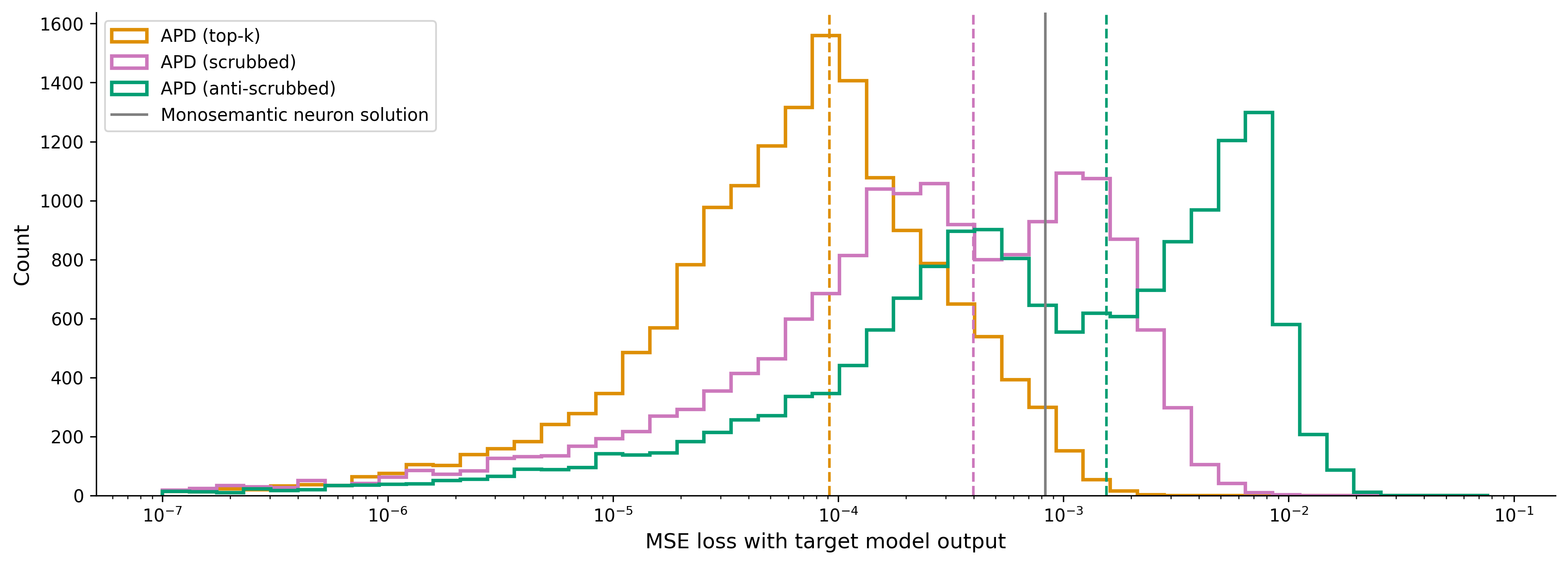}
    \caption{MSE losses of the $2$-layer residual MLP APD model on the sparse forward pass (``top-$k$") and the APD model when ablating half ($50$) of its parameter components (``scrubbed" when none of the components responsible for the active inputs are ablated and ``anti-scrubbed" when they are ablated). The gray line indicates the loss for a model which uses one monosemantic neuron per input feature.}
    \label{fig:scrubbing-2layers}
\end{figure}

To further show that APD is indifferent to computations occurring in multiple layers, we replicate the $1$-layer figures (\ref{fig:weight-linearity} and \ref{fig:scrubbing}) for the $2$-layer setting in Figures \ref{fig:weight-linearity-2layers} and \ref{fig:scrubbing-2layers}, respectively. The qualitatively similar results indicate that despite the learned parameter components representing computation occurring across multiple layers, the components have minimal influence on forward passes when their corresponding input feature is not active.

\section{Training details and hyperparameters}\label{app:training-details}

\tocless\subsection{Toy models of superposition (TMS)}\label{app:training-details-tms}

\tocless\subsubsection{TMS with $5$ input features and hidden size of $2$}
The target model was trained for $5$k steps with a batch size of $1024$. We use the AdamW optimizer \citep{loshchilov2019adamw} with a weight decay of $0.01$ and a constant learning rate of $0.005$. Our datasets consists of samples with each of the $5$ input features taking values in the range $[0,1]$ (uniformly) with probability $0.05$ and $0$ otherwise.

To train the APD model for TMS, we use the Adam optimizer \citep{kingma2017adam} with a constant learning rate of $0.03$ with a linear warmup over the first $5\%$ steps. We use the same data distribution as for training the target model (feature probability $0.05$). We train for $20$k steps with a batch size of $2048$, and a batch top-k value of $0.211$, indicating that $0.211\times2048=432$ parameter components are active in each batch. The coefficients for the loss functions are set to $1$ for $\mathcal{L}_{\text{faithfulness}}$, $1$ for $\mathcal{L}_{\text{minimality}}$, and $0.7$ for $\mathcal{L}_{\text{simplicity}}$ with a $L_p$ norm of $1$.

\tocless\subsubsection{TMS with $40$ input features and hidden size of $10$}
The target model was trained for $2$k steps with a batch size of $2048$ (we expect we would have achieved the same results with $5$k steps and batch size $1024$, as we used for the smaller TMS setting). We use AdamW with a weight decay of $0.01$ and learning rate constant learning rate of $0.005$. Our datasets consists of samples with each of the $5$ input features taking values in the range $[0,1]$ (uniformly) with probability $0.05$ and $0$ otherwise.

To train the APD model, we use Adam with a max learning rate of $0.001$ that decays with a cosine schedule and has a linear warmup over the first $5\%$ steps. We use $50$ components, allowing for $10$ to `die' throughout training. We use the same data distribution as for training the target model (feature probability $0.05$). We train for $20$k steps with a batch size of $2048$, and a batch top-k value of $1$, indicating that an average of $1\times2048=2048$ parameter components are active in each batch. The coefficients for the loss functions are set to $1$ for $\mathcal{L}_{\text{faithfulness}}$, $10$ for $\mathcal{L}_{\text{minimality}}$, and $20$ for $\mathcal{L}_{\text{simplicity}}$ with a $L_p$ norm of $0.9$.

\tocless\subsection{Compressed computation and cross-layer distributed representation}\label{app:training-details-resid-mlp}
Recall that the $1$-layer residual MLP (Section \ref{sec:residmlp_1layer}) and $2$-layer residual MLP (Section \ref{sec:residmlp_2layer}) both have $100$ input features, an embedding dimension of $1000$, and $50$ MLP neurons ($25$ in each MLP layer for the $2$-layer case). Both target models were trained using AdamW with a weight decay of $0.01$, a max learning rate of $0.003$ with cosine decay, batch size of $2048$. The datasets consist of samples with each of the $100$ input features taking values in the range $[-1,1]$ (uniformly) with probability $0.01$ and $0$ otherwise.

Both $1$-layer and $2$-layer APD models were trained with the Adam optimizer with a max learning rate of $0.001$ which had a linear warmup for the first $1\%$ of steps and a cosine decay thereafter. The models were trained with a batch size of $2048$, and a batch top-k value of $1.28$, indicating that $1.28\times2048=2621$ parameter components are active in each batch. Both models have a coefficient set to $1$ for $\mathcal{L}_{\text{faithfulness}}$, and $1$ for a loss which reconstructs the activations after the non-linearity in the MLP layers.

The $1$-layer model starts with $130$ parameter components, trains for $40$k steps, has a coefficient of $1$ for $\mathcal{L}_{\text{minimality}}$ and $10$ for $\mathcal{L}_{\text{simplicity}}$ with a $L_p$ norm of $0.9$. We also apply a normalization to the factorized form of the parameter components. Specifically, we normalize $U$ in Equation \ref{eq:factorized} every training step so that it has unit norm in the in\_dim dimension (labeled $i$ in the equation). We expect that it's possible to achieve equivalent performance and stability without this normalization with a different set of hyperparameters.

The $2$-layer model starts with $200$ parameter components, trains for $10$k steps, has a coefficient of $2$ for $\mathcal{L}_{\text{minimality}}$ and $7$ for $\mathcal{L}_{\text{simplicity}}$ with a $L_p$ norm of $0.9$.

We note that many of the inconsistencies in hyperparameters between different experiments are not due to rigorous ablation studies, and we expect to obtain similar results with more consolidated settings. In particular, changes to learning rate configurations (warmup, decay), training steps, $L_p$ norm for $\mathcal{L}_{\text{simplicity}}$, and batch size, did not tend to have a large influence on the results. Other hyperparameters such as the coefficients of the loss terms, and, to a lesser extent, the batch top-k value, do have a significant influence on the outcome of APD runs.

\tocless\subsection{Hand-coded gated function model}

Recall that our experiments use $4$ unique functions, with each function using $m=10$ neurons to piecewise-approximate each of the $4$ functions.

For APD training, we use Adam with a max learning rate of $0.003$ that decays with a cosine schedule and has a linear warmup over the first $0.4\%$ steps. Our dataset consists of one input variable whose entries are drawn uniformly from $[0,5]$ and four control bits taking a value of $1$ with probability $0.05$ and $0$ otherwise. We train for $200$k steps with a batch size of $10000$, and a batch top-k value of $0.2217$, indicating that $0.2217\times10000=2217$ parameter components are active in each batch. The coefficients for the loss functions are set to $0.1$ for $\mathcal{L}_{\text{faithfulness}}$, $5$ for $\mathcal{L}_{\text{minimality}}$, and $5$ for $\mathcal{L}_{\text{simplicity}}$ with a $L_p$ norm of $1$.